\pdfoutput=1
\documentclass[review]{elsarticle}
\usepackage{subfigure}
\usepackage{longtable}
\usepackage{algorithm, algorithmic}
\usepackage{bm}
\usepackage{booktabs}
\usepackage{multirow}
\usepackage{geometry}
\usepackage{amsfonts,amssymb}

\usepackage{lineno,hyperref}
\modulolinenumbers[5]

\journal{arXiv}

\begin{document}

\begin{frontmatter}

\title{Adaptive Online Incremental Learning for Evolving Data Streams}


\author[mymainaddress]{Si-si Zhang}

\author[mymainaddress]{Jian-wei Liu\corref{mycorrespondingauthor}}
\cortext[mycorrespondingauthor]{Corresponding author}
\ead{liujw@cup.edu.cn}

\author[mymainaddress]{Xin Zuo}

\address[mymainaddress]{Department of Automation, College of Information Science and Engineering,
China University of Petroleum , Beijing, Beijing, China}

\begin{abstract}
Recent years have witnessed growing interests in online incremental learning. However, there are three major challenges in this area. The first major difficulty is concept drift, that is, the probability distribution in the streaming data would change as the data arrives. The second major difficulty is catastrophic forgetting, that is, forgetting what we have learned before when learning new knowledge. The last one we often ignore is the learning of the latent representation. Only good latent representation can improve the prediction accuracy of the model. Our research builds on this observation and attempts to overcome these difficulties. To this end, we propose an Adaptive Online Incremental Learning for evolving data streams (AOIL). We use auto-encoder with the memory module, on the one hand, we obtained the latent features of the input, on the other hand, according to the reconstruction loss of the auto-encoder with memory module, we could successfully detect the existence of concept drift and trigger the update mechanism, adjust the model parameters in time. In addition, we divide features, which are derived from the activation of the hidden layers, into two parts, which are used to extract the common and private features respectively. By means of this approach, the model could learn the private features of the new coming instances, but do not forget what we have learned in the past (shared features), which reduces the occurrence of catastrophic forgetting. At the same time, to get the fusion feature vector we use the self-attention mechanism to effectively fuse the extracted features, which further improved the latent representation learning. Moreover, in order to further improve the robustness of the algorithm, we add the de-noising auto-encoder to original framework. Finally, we conduct extensive experiments on different datasets, and show that the proposed AOIL gets promising results and outperforms other state-of-the-art methods. 
\end{abstract}


\end{frontmatter}

\section{Introduction}

In machine learning, the typical learning approaches are to divide dataset into training set, cross validation set and test set. When the model completes the training process on the training set and we determines the hyper-parameters of the model on the cross validation set, then we would directly use them to evaluate the prediction performance on the test set. This approach is based on the fact that the probabilistic distribution data obey would not change. However, in practice, the input data is processed in the form of stream. When the sequentially arrived samples deviate from the original probability distribution followed by the previous samples, or when the model cannot get enough training sets, the model tends to misclassify these samples. In order to address this issue, online incremental learning, also known as continuous learning or lifelong learning  \cite{1elwell2011incremental,2brzezinski2014combining,3ristin2015incremental,4krawczyk2018online,5ren2017life} has emerged. Online incremental learning, as an alternative learning paradigm, is a more natural and humanized learning paradigm. Its aim is to overcome some of the limitations above.

The goal of such algorithms is to constantly update model parameters according to incoming stream data and constantly learn new knowledge. As is known to all, data streams are usually divided into stationary and non-stationary data streams. When dealing with non-stationary data streams, it means that samples may come from different probability distributions, i.e., concept drift problems occur. Specifically, there may be variations in the probability distribution, i.e., at two consecutive moments $t$ and $t-1$,$P(X_t,Y_t) \neq P(X_{t-1},Y_{t-1})$.

There are two main methods to deal with concept drift, one is active attack, and the other is passive update\cite {6lobo2020spiking}. The former method needs to detect whether there is concept drift in streaming data, once the concept drift is detected, then trigger update mechanism. The other is passive update, that is, when each instance comes, it needs to update model parameters continually. Because it is easy to implement, and it does not need to consider when to update, the most of the online incremental learning approaches adopt the second approach, which also called blind approaches. For the first method, there are also some state-of-art learning methods, such as Adwin algorithm, which was introduced by Albert Bifet \cite{7bifet2007learning}.It presents a sliding window with variable size to calculate the size of the window online according to the rate of change observed from the data of the window itself. This method could automatically increase the size of window when there is no obvious change, and reduce the size of window when the data constantly occurring changes.

To make progress in this direction, this paper presents a novel approach to detect the concept drift, which calculates the total loss directly. Obviously, if the probability distribution of the data changes and the model was not updated in time, the prediction loss of the algorithm would increase. Checking for concept drift by examining changes in predicted losses is commonly used by many algorithms \cite{8fowler2011anomaly}. Apparently, this is retrospective wisdom, which cannot adjust the model in time. However, the key idea of our proposed approach is how to deal with the problem after concept drift is found. In this paper, we add the reconstruction loss to the original prediction loss, and hope to learn a model being capable of detecting and updating the model in time when the concept drift occurs. Thus, we build an auto-encoder to detect whether the concept drift has occurred according to the reconstruction error. When the probability distribution of the input data is different from the reconstructed one by auto-encoder we trained before, the reconstruction loss would be greatly increased. According to this point, the occurrence of concept drift is detected. However, if the incoming streaming data existing concept drift shares common compositional patterns with the previous training data or the decoder is ‘too strong’, which can decode the data existing concept drift well. The encoder seems to have lost the ability to detect concept drift. Hence, we adapt the auto-encoder with memory module to overcome this problem \cite{9gong2019memorizing}.

Given streaming data, we want to learn them one by one to accumulate new knowledge from the new one, but sometimes we may forget what we have learned before. This is another problem we encounter in online learning called catastrophic forgetting. We need to make a trade-off between the plasticity and predictive accuracy \cite{10kirkpatrick2017overcoming}.Another classic saying is making a good compromise between exploration and exploitation \cite{11deckert2013incremental},where exploration means aggressively searching new knowledge, exploitation means making full use of experience to help current learning. The most classical algorithm for catastrophic forgetting was introduced by Kirkpatrick, who proposed an approach called elastic weight combination (EWC)\cite{12wang2017feature}. All these types of algorithms start from the output layer, that is, focus on control the whole network through the objective function, and let the model weigh and consider the balance between these two factors, rather than directly take measures to deal with the problem of catastrophic forgetting. However, these types of methods disregard the underlying correlations between different features and ignore the latent representation learning. In this regard, therefore, this reminds us of the importance of strengthening these two aspects.

In the online incremental learning arena, the existing network structure cannot utilize well the information extracting from different latent layers of the deep network. To promote prediction performance, it is very important to learn a good latent representation from different abstract level. In fact, the auto-encoder and its variants show promising results in extracting effective features. As we have introduced the auto-encoder in the concept drift detection part, we only need to take the hidden layers of the encoder to extract the input information effectively. 

After realizing the learning of the hidden layer, what we need to do is to reduce the occurrence of catastrophic forgetting. The reason for this problem is that the model only focuses on learning new knowledge and forgets what it has learned in the past. Hence, in this paper, we try to get information extracting from different latent layers of auto-encoder by jointly exploring the both properties of common and private features on streaming data. The specific implementation is to divide the hidden layer of auto-encoder into common features and private features artificially. The common features are used to learn the shared features on streaming data, and the private features are used to learn the proprietary properties of new instances on streaming data. In addition, to take full advantage of the feature set of these different attributes, a self-attention mechanism is designed to effectively fuse these features to get better final feature vectors.

Another issue we think needs to be considered is the robustness of the algorithm. Streaming data would more or less contain noise, but there are few specific measures to deal with this problem in the current researches. Therefore, when there is a large amount of additional meaningless noise in the data, we incorporate the de-noising auto-encoder to improve the robustness of the algorithm.

The contributions of this paper are as follows:
 
1) Constructing an auto-encoder with a memory module in an online incremental learning scenario can immediately detect concept drift and trigger the model update mechanism.

2) The hidden representations for different abstract level are considered and the self-attention mechanism is used to extract and fuse these hidden representations effectively so as to obtain a compact and comprehensive hidden representation.

3) In terms of feature level, our method effectively extracts the shared and private features on the streaming data and learns new knowledge without forgetting what have learned in the past.

4) To handle the noisy streaming data, de-noising auto-encoder is considered in our algorithm. In addition, the effectiveness of the algorithm is verified on different datasets.

Our paper is organized as follows. In section 2, we introduce some related works on online   learning. In section 3, we described the adaptive online incremental learning for evolving data streams. In section 4, the specific implementation of proposed AOIL algorithm is proposed, and we summarize the results of previous discussions. In section 5, the performance is evaluated and compared to the state of the art models in the environment of streaming data. In section 6, we will make conclusion for this paper.

\section{Related work}

\subsection{Online incremental learning}

In the traditional offline learning, all the training data must be available for the algorithm before learning, however, in practice, we often can't get all the training sets. The data is in the form of streaming data, and the model must have the ability of real time processing. As a result, the concept of online incremental learning has emerged, which optimizes the model parameters on the data stream in order to improve the performance of the algorithm. And it has achieved a lot of success both in application and theory research \cite{13cesa2021online,14bae2017confidence,15de2015evolutionary,16precup2020evolving}.However, the previous online incremental models are based on linear function or kernel function for nonlinear problem, but choosing the appropriate kernel function is also a problem to be solved. With the development of deep learning, the role of machine learning algorithms becomes more prominent. However, when online incremental learning is applied directly to the deep learning with online back propagation, there are some shortcomings such as difficulty in training (gradient vanishing and feature reuse decreasing). Zhou et al. propose an algorithm can effectively identify the change of probability distribution in streaming data based on de-noising auto-encoders \cite{17zhou2012online}.Lee et al. propose a dual memory architecture, which can not only deal with the slow changing global pattern, but also track the sudden and rapid changes locally \cite{18lee2016dual}.These are some algorithms for fixed model and some variable structure model, such as Sahoo et al. propose an online incremental learning framework from shallow to deep \cite{19sahoo2017online};Ashfahani et al introduce a different-depth structure to handle data streaming \cite{20ashfahani2019autonomous}. However, they all ignore the problem of catastrophic forgetting, or they cannot deal with it very well. It is impossible to get good results by adjusting the model structure without fully analyzing the characteristics of samples.

\subsection{Data stream learning}

Generally, data types in streaming data are divided into stationary and non-stationary data stream. Non-stationary data stream refers to the phenomenon that probability distribution in streaming data would change with the incremental arrival of the streaming data, e.g., the problem of concept drift. At present, there are two types of strategies to deal with concept drift as just mentioned. One is to update the model as long as the new data comes, no matter whether there is concept drift or not. Another is to take measures to detect if there is a concept drift, and if there is, trigger an update mechanism to accommodate the new probability distribution. Most of the existing models use the first strategy \cite{21gama2004learning}. As a result of constantly updating their model, it makes it evolve in a very regular way regardless of the concept drift. 

Among online incremental learning algorithms with the first strategy, the following can be mentioned: Approximate Maximal Margin Classification algorithm (ALMA) \cite{22gentile2001new}; the Relaxed Online Maximum Margin Algorithm and its aggressive version aROMMA, ROMMA, and aROMMA \cite{23li2002relaxed}; the Adaptive Regularization of Weight Vectors(aROW) \cite{24crammer2009adaptive}; the Confidence-Weighted (CW) learning algorithm, \cite{25crammer2008exact}; new variant of Adaptive Regularization(NAROW) \cite{26orabona2010new}; the Normal Herding method via Gaussian Herding (NHERD) \cite{27crammer2010learning}; the Online Gradient Descent (OGD) algorithms \cite{28zinkevich2003online}; the recently proposed Soft Confidence Weighted algorithms (SCW) \cite{29wang2012exact}; second-order perception (SOP) \cite{30cesa2005second}.

In addition, there are some learning methods based on the detection of data probability distribution changes, including: A concept drift method, using adaptive online sliding windows according to the rate of change observed from the data(Adwin) \cite{7bifet2007learning}, which can be classified as windows based methods; PageHinkley \cite{31gama2013evaluating}, which could be see as a sequential analysis based methods \cite{32gama2014survey}. When the algorithm detects that the new sample deviates from its average value, it is judged as a concept drift, and a drift alarm is given when the difference exceeds the preassigned threshold value.

\section{Adaptive Online Incremental Learning for Evolving Data Streams}

In this section, we would embark on a discussion of the proposed Adaptive Online Incremental Learning for evolving data streams (AOIL). Firstly, we would recall online learning scenario briefly, and then proposes our framework. AOIL algorithm would introduced in five parts: latent representation learning, auto-encoder with memory module, the self-attention mechanism for feature fusion of hidden layer, the proposed objective function and adaptive dynamic update for drifted parameters of network.

In the latent representation learning, we reconstruct the input using auto-encoder and acquire the latent feature representation. In order to better detect the concept drift, we hope that when the concept drift occurs, the auto-encoder’s reconstruction loss on streaming data is quite different from the previously one. However, the normal auto-encoder cannot distinguish the concept drift very well, or the reconstruction loss of the normal auto-encoder for the two streaming data following different probability distributions is not much different. Therefore, the auto-encoder with memory module is suitable for simultaneously implementing feature coding and concept drift detection. After obtaining the latent representation, we present the feature fusion strategies by self-attention mechanism and consider the reconstruction loss and prediction loss, deriving the whole optimization objective function. Finally, in the section of adaptive dynamic update for drifted parameters of network, we would specifically introduce the detection of concept drift and relieve the catastrophic forgetting problem. 

\subsection{Learning Setup}

In an online incremental learning scenario, learning on evolving data streams means learning the unknown distribution of streaming data. More specifically, the main goal of constructing a classification on data streams is to build an incremental model to predict the label of input data according to the available information, which requires less memory space. Learning setup on data streams can be defined as follows:

Let $S$ be data streams, which are composed of sequences $(x_1,y_1),(x_2,y_2),...,(x_t,y_t)$.At each time instant $t$ the agent acquires the example $x_t = (x_{t,1},x_{t,2},...,x_{t,D_x})^T \in R^{D_x}$,which is derived from the current probability distribution $D_t$.However, the current distribution $D_t$ may different from previous probability distribution of $D_{t-1}$.. The agent does not know when this change would happen and was required to predict its output $y_t \in R^{D_\gamma}$ through learning a mapping $f(\cdot)$ according to information of the previous input-output sequence   $(x_1,y_1),(x_2,y_2),...,(x_{t-1},y_{t-1})$.Due to the limitation of storage capacity and high throughput arrival streaming data, storing previous training data is forbidden. After obtaining the prediction value $\hat{y}_t \in R^{D_\gamma}$ from the network, the agent would receive the real output value $y_t \in R^{D_\gamma}$ from the environments. By calculating the loss between the prediction and the ground-truth value, the current model of agent suffers the prediction loss. Then the loss information would feed back to the learning algorithm of agent to guide the updating process of the model parameters.  

\subsection{Latent Representation Learning}

The success of machine learning algorithms generally depends on how to learn a good latent representation on data, and we guess that this is because different representations can entangle more or less the different explanatory factors of variation on data, and may help infer the hidden underlying knowledge of the data \cite{33bengio2013representation}. Therefore, it is important to learn latent representations of the input that make it easier to extract useful information when building classifiers or other predictors. However, there is little research in the field of online incremental learning on how to use a deep network to learn the latent representation and realize feature selection, and then adopt the hidden representation to predict class labels of examples. Hence, we utilize auto-encoder to extract abstract hierarchical features from example, yield input example’s hidden representation. When the parameters of the auto-encoder are effectively trained, the auto-encoder can become a powerful generation model and an effective representation learning tool for future prediction.

The architecture flowchart of the auto-encoder is depicted in Fig. 1. The left panel in Fig. 1 is encoder and the right panel in Fig. 1 is decoder. The encoder receives the input sequentially and builds a fixed-length latent vector representation (denoted as $h_5 \in \mathbb{R}^{D_h}$ in Fig.1). Conditioned on the encoded latent representation, the decoder generates the reconstructed input (denoted as $\hat{x} \in \mathbb{R}^{D_x}$ in Fig.1). 
\begin{figure}[!htbp]
	\centering
	\includegraphics[scale=0.7]{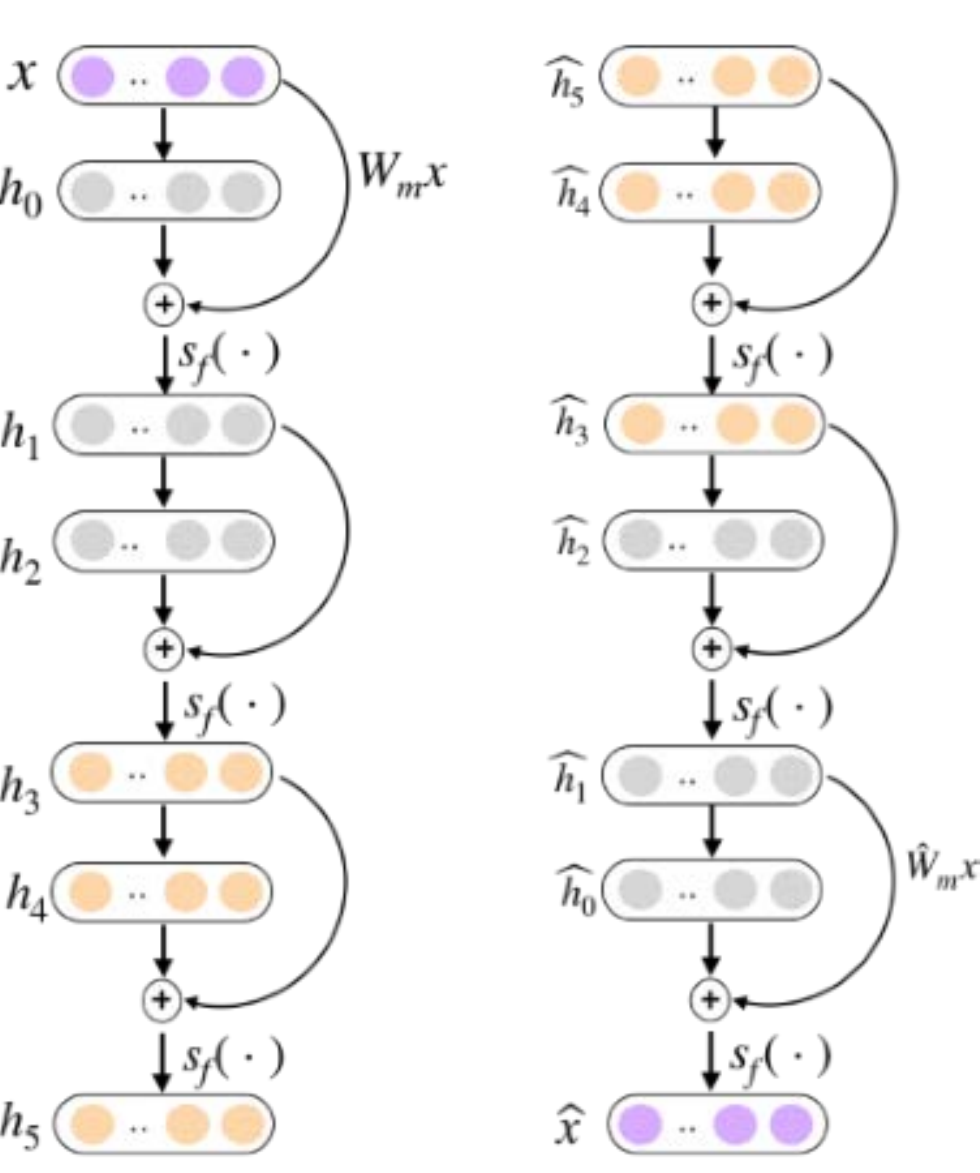}
	\caption{The architecture of the encoder of auto-encoder with ResNet network}
	\label{fig1}
\end{figure}

In order to extract more information from input examples, and make significant improvements for prediction accuracy, we increase the depth of the hidden layer up to 6 layers. However, if we simply increase the depth, there may occur degradation problems that seriously affect the predictive performance. Hence, we introduce the architecture of the ResNet network \cite{34he2016deep}, a solution to address this problem, allowing us to efficiently learn much deeper networks. This is accomplished by using a shortcut connection, in our case, we chose to skip one hidden layer. Shortcut connections $[h_0,h_2,h_4,\hat{h_0},\hat{h_2},\hat{h_4}]$ are those skipped one.

Hence, the transformation can be formulated as follows:
\begin{equation}
\label{eq1}
\left\{ {\begin{array}{*{20}c}
	{ {h_0 = f_\theta (x) = s_f(b_0+W_0x) } ,}  \\
	{ {h_1 = s_f(b_1+W_1h_0+W_mx) } ,}  \\
	{ {h_2 = s_f(b_2+W_2h_1) } ,}  \\
	{ {h_3 = s_f(b_3+W_2h_1) } ,}  \\
	{ {h_4 = s_f(b_4+W_4h_3) } ,}  \\
	{ {h_5 = s_f(b_5+W_5h_4+h_3) } .}  \\
	\end{array}} \right.	
\end{equation}
\begin{equation}
\label{eq2}
\left\{ {\begin{array}{*{20}c}
	{ {\hat{h_4} = s_f(\hat{b_5}+\hat{W_5}\hat{h_5}) } ,}  \\
	{ {\hat{h_3} = s_f(\hat{b_4}+\hat{W_4}\hat{h_4}+\hat{h_5}) } ,}  \\
	{ {\hat{h_2} = s_f(\hat{b_3}+\hat{W_3}\hat{h_3}) } ,}  \\
	{ {\hat{h_1} = s_f(\hat{b_2}+\hat{W_2}\hat{h_2}+\hat{h+3}) } ,}  \\
	{ {\hat{h_0} = s_f(\hat{b_1}+\hat{W_1}\hat{h_1}) } ,}  \\
	{ {\hat{x} = s_f(\hat{b_0}+\hat{W_0}\hat{h_0}+\hat{W_m}\hat{h_1}) } .}  \\
	\end{array}} \right.	
\end{equation}

Where $s_f(\cdot)$ is Relu activation function of the encoder and decoder, the set of parameters for such an auto-encoder is generally defined as ${\theta}_l = \left\{ W_l,b_l,\hat{W_l},\hat{b_l} \right\},l=0,1,...,5$.Here $W_l$,$b_l$ and $\hat{W_l}$,$\hat{b_l}$ are the weight matrices and bias vectors of encoder and decoder respectively. Where $W_m \in \mathbb{R}^{D_h \times D_x}$ and $W_m^{\prime} \in \mathbb{R}^{D_x \times D_h}$ are matrices introduced to match dimensions between different layers.

In addition, the deviation between the input and reconstructed one was described by mean square error, which is specifically defined as $L_{re}(x,\hat{x})=\Vert x\Vert _2^2$.

\subsection{Auto-encoder with Memory Module}

Auto-encoders are capable of taking a high dimensional vector of the instance space as inputs and map them into the latent space of reducing dimensionality that captures the critical hidden information of the inputs. When the network parameters are trained with the streaming data, the difference between the input and reconstructed one becomes large when the current input data is significantly different from the previously trained pattern. So Auto-encoders are extensively used for unsupervised anomaly detection \cite{35nolle2018analyzing,36ellefsen2020online}. In our case, we tend to use encoder as our means to detect concept drift. However, if the incoming streaming data exists concept drift which shares common compositional patterns with the previous training data or the decoder is ‘too strong’, which can decode the existing concept drift in data well, the encoder seems to lose the function of detecting concept drift.
\begin{figure}[!htbp]
	\centering
	\includegraphics[scale=0.7]{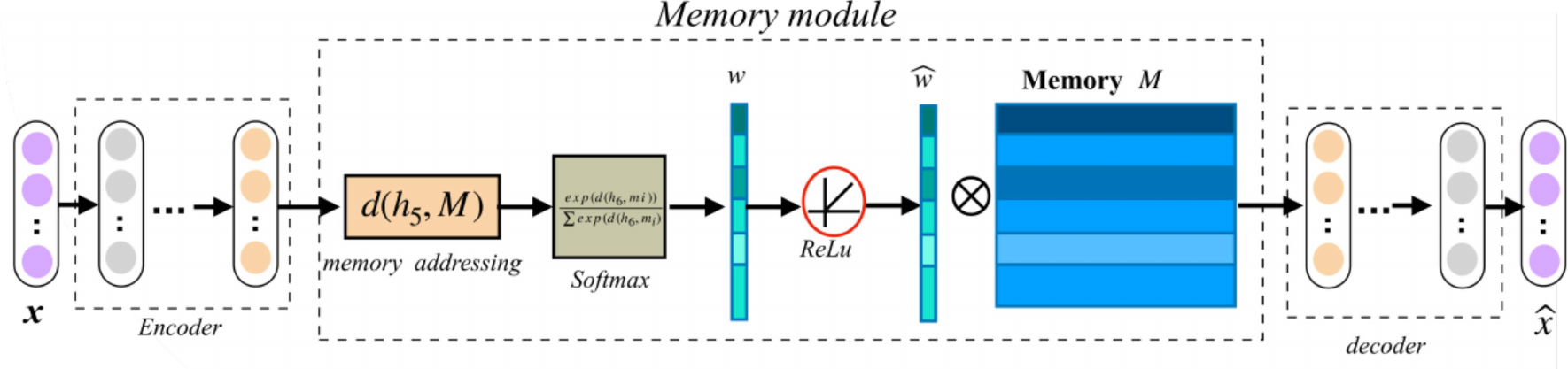}
	\caption{Autoencoder with Memory Module}
	\label{fig2}
\end{figure}

In order to circumvent this issue, we incorporate an encoder with memory module \cite{9gong2019memorizing}, the specific structure are illustrated in the Fig.2. The memory module is designed as a memory matrix $M \in \mathbb{R}^{N \times D_h}$,recording the prototypical stationary patterns in data streams, it represents $N$ memory units $m_i,i=1,...,N$,with fixed size  $D_h$,dimension of each memory unit is same as the hidden layer in encoder.

Specifically, we obtain the decoder layer $\hat{h}\in \mathbb{R}^{1 \times D_h}$ by a non-negative soft addressing vector $w\in \mathbb{R}^{1 \times N}$:
\begin{equation}
\label{eq3}
\hat{h_5}=\hat{w}M=\sum{_{i=1}^N}\hat{w_i}m_i
\end{equation}
To ensure that the elements in $\hat{w}$ are not-negative, we use $w \in \mathbb{R}^{1\times N}$ as input of the ReLu function to build the elements of $\hat{w}$.
\begin{equation}
\label{eq4}
\hat{w_i}=\frac{max(w_i,0)\cdot w_i}{\mid{w_i}\mid+\varepsilon}
\end{equation}

Where the components of $w$ are derived from outputs of a softmax function with memory units $m_i$ and the activation of last layer of encoder $h_5\in\mathbb{R}^{1\times D_h}$ as inputs:
\begin{equation}
\label{eq5}
w_i=\frac{exp(d(h_5,m_i))}{\sum{_{j=1}^N}exp(d(h_5,m_j))}
\end{equation}

Where $d(\cdot ,\cdot)$ denotes the metric for cosine similarity:
\begin{equation}
\label{eq6}
d(h_5,m_i)=\frac{h_5m_i^T}{\Vert h_5\Vert\Vert m_i\Vert}
\end{equation}

The use of memory modules can be explained as follows: in the stationary streaming data, due to the restricted memory size and the sparse memory addressing operation in Eq.(4), It enables the network to make full use of memory units, records the most representative features of the stationary data. The memory unit can remember the previous data pattern. When the concept drift occurs, although the pattern of input data has changed, the memory unit still keeps the state that the probability distribution of data has not changed. Therefore, the reconstruction loss of the auto-encoder would be great, which is convenient for us to find this change and adjust the parameters in time.  
In order to promote the sparsity of $\hat{w}\in\mathbb{R}^{1\times N}$ and give full play to the role of the memory module, a new loss function of the au-encoder is constructed by applying a regularization term to the memory module.
\begin{equation}
\label{eq7}
L_{re}=L(x,\hat{x})+\lambda L(\hat{w})
\end{equation}

Where $L(\hat{w})=\sum{_{i=1}^N}-\hat{w_i}log(\hat{w_i})$ is the regularized item which is related to memory module, and $\lambda$ is the trade-off hyper-parameter. 

Accordingly, we regard the loss $L_{re}$ as an important part of the total loss, to judge whether there is concept drift in the streaming data.

\subsection{The Self-Attention Mechanism for Feature Fusion of Hidden Layer}

After extracting features from the hidden layer of the encoder, we notice that not all features contribute the same to the classifier. Therefore, it is not conducive to deal with the continuous evolution relationship between feature selection by evenly distributing the weights of the hidden layer or defining the weights as constants in advance.

In order to handle the relationships between feature and label with evolving complicated distribution, the self-attention mechanism \cite{37lin2017structured} is used to provide a set of summative weight vectors for the hidden layer of the encoder, assign those more important features by giving them a higher weight to increase their importance.
\begin{figure}[!htbp]
	\centering
	\includegraphics[scale=0.7]{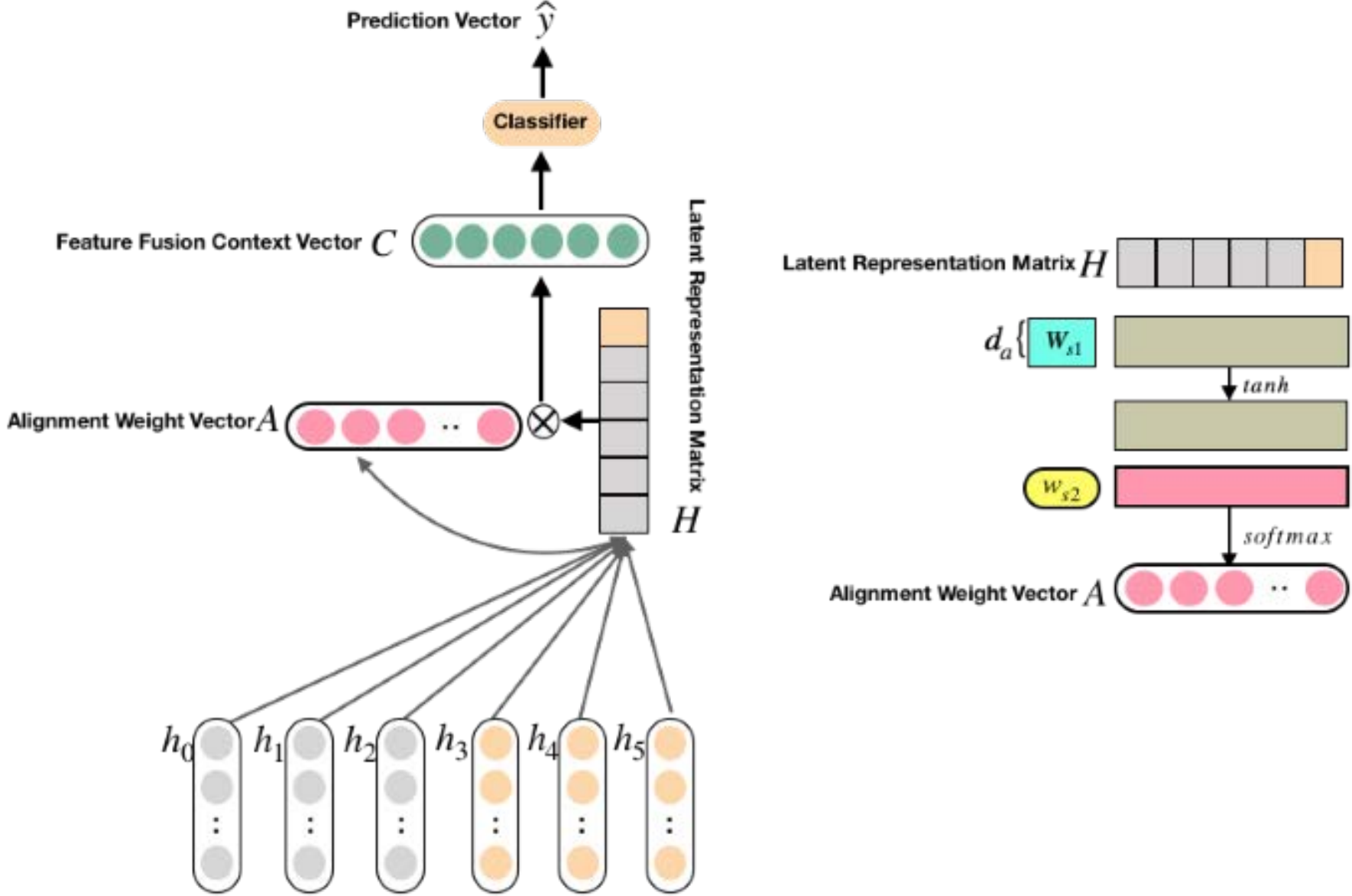}
	\caption{The left panel is the Self-Attention Mechanism for feature fusion of hidden layer, the right panel is the process of getting alignment weight vector A.}
	\label{fig3}
\end{figure}

From the Fig.3 we can see the feature fusion process of AOIL. Instead of directly constructing the classifiers from the outputs of the hidden layers, we first fusion the outputs of the hidden layers to form context latent representation. At each time, we first concatenate them to form a latent representation matrix  $H=(h_0,h_1,...h_L)$, where $H$ is a $(L+1)\times D_h$ matrix, as illustrated by Fig.3. The self-attention mechnisim use $H$ as input and obtain the alignment weight vector $A=[a_0,a_1,...a_L]$ through a softmax layer: 
\begin{equation}
\label{eq8}
A=softmax(w_{s2}tanh(W_{s1}H^T))
\end{equation}

Where $W_{s1}\in\mathbb{R}^{d_a\times D_h}$ is a weight matrix, and $w_{s2}\in\mathbb{R}^{d_a}$, the dimension $d_a$ is predefined hyperparameter.$A\in\mathbb{L+1}$ is a alignment weight vector, which represent the contributions of the columns $h_i$ of $H$.Due to the role of softmax, it ensures that the sum of components in vector $A$ equals to 1. 

Meanwhile, the result of feature fusion, i.e., the context vector $C$ can be obtained by multiplying the alignment weight vector and the latent representation matrix $H$:
\begin{equation}
\label{eq9}
C=AH
\end{equation}

Finally, the results of feature fusion are fed into the output layer of classifier:
\begin{equation}
\label{eq10}
\hat{y_t}=softmax(C\cdot W_f+b_f)
\end{equation}

In this way, we get our prediction label $\hat{y_t}$. Moreover, the cross-entropy metric is adopted to get the prediction loss $L_{pre}(y_t,\hat{y_t})=-\sum y_tlog(\hat{y_t})$.

\subsection{The Proposed Objective Function}

We now have obtained the prediction output according to the feature fusion strategy and the autoencoder with memory module. The reconstruction loss of the auto-encoder with memory moudle is defined as $L_{re}=L(x,\hat{x})+\lambda L(\hat{w})$.In addition, the prediction loss is defined as cross-entropy loss $L_{pre}(y_t,\hat{y_t})=-\sum y_tlog(\hat{y_t})$, which represents the distance between the actual probability and the prediction probability of label. Combining these two losses, we can formulate our total objective function of the online learning system as follows:
\begin{equation}
\label{eq11}
L_{total}=L_{pre}(y,\hat{y})+L(x,\hat{x})+\lambda L(\hat{w})
\end{equation}

Apparently, the motivation for this loss is to explicitly balance the reconstruction loss and prediction loss and optimize them jointly.

In addition, different from previous traditional method, here, for updating the parameters of the autoencoder $\theta_l=\left\{ W_l,b_l,\hat{W_l},\hat{b_l}\right\},l=0,1,...,5$, the loss of back-propagation is composed of two parts, reconstruction loss $L(x,\hat{x})+\lambda L(\hat{w})$ and prediction loss $L_{pre}(y,\hat{y})$.

However, the back propagation for prediction loss is more involved. For learning each parameter of the encoder’s hidden layer, we use the sum of the gradient of the prediction loss on current hidden layer and its subsequent hidden layers, i.e.,$\sum{_{j=l}^L}\nabla_{\theta_l^{t-1}}L_{pre}(y_{t-1},\hat{y_{t-1}})$.

Meanwhile, in order to further play the role of self-attention mechanism, we add the attention alignment weight to $\sum{_{j=l}^L}\nabla_{\theta_l^{t-1}}L_{pre}(y_{t-1},\hat{y_{t-1}})$,thus the final gradient related to prediction loss has form $\sum{_{j=l}^L}\alpha_j\nabla_{\theta_l^{t-1}}L_{pre}(y_{t-1},\hat{y_{t-1}})$, plus the back propagation of reconstruction loss, the whole update formula for the parameters of the autoencoder is as follows:
\begin{equation}
\label{eq12}
\theta_l^t\leftarrow \theta_l^{t-1}-\delta(\sum{_{j=l}^L}\alpha_j\nabla_{\theta_l^{t-1}}L_{pre}(y_{t-1},\hat{y_{t-1}})+\nabla_{\theta_l^{t-1}}L(x_{t-1},\hat{x_{t-1}})+\lambda \nabla_{\theta_l^{t-1}}L(\hat{w_{t-1}}))
\end{equation}

where $\delta$ represents the learning rate. $\sum{_{j=l}^L}\alpha_j\nabla_{\theta_l^{t-1}}L_{pre}(y_{t-1},\hat{y_{t-1}})$ is computed via derivatives of backpropagation error of $\hat{y_{t-1}}$.Note that where the index of summation starts at $j=l$ end with $j=L$ and because the backpropagation error of shallower layer depend on all the deeper layers. In fact, we derive the final gradient with respect to the backpropagation derivatives of a predictor at every hidden layer ,weighted by alignment weight $\alpha_j$ of self-attention mechanism.

\subsection{Adaptive Dynamic Update Network }

In this subsection, we would describe how to detect concept drift and how to dynamically update parameters to prevent catastrophic forgetting. When concept drift occurs, it is obvious that the network structure needs to be adjusted to accommodate concept drift on the data streams. However, existing online continuous learning methods often do this by increasing or decreasing the depth and width of the network \cite{38pratama2019atl,39ashfahani2020devdan,40pratama2019incremental,41yoon2017lifelong}. This approach usually increases the complexity of the network, and we know that when the network depth reaches a certain level, increasing the depth of the network would not improve the effectiveness of the algorithm. In addition, one neglected fact about streaming data is that, although there is concept drift in the streaming data, there is still some association between the current examples and previous examples, that is, the current examples in the streaming data has not only private features, but also shared features with previous examples. Although the private features achieve a more compact representation for a single current example, the shared features for the whole set are able to provide more efficient and robust representations when the final model is trained to fit the whole streaming data. Based on this idea, as examples in streaming data tend to have many shared properties, the effective latent representation learning approach should capture those commonalities. Therefore, based on feature extraction, we present a learning procedure to reduce computational complexity by finding common features that can be shared across streams of data. Clearly, unlike existing traditional approaches, we do not fix the parameters of the hidden layer of the encoder. The idea is to use the first three layers of the auto-encoder to preserve the shared features of the data stream, while the last three layers are used to extract the private features. 

\subsubsection{The Detection of Concept Drift}

Recall that we can judge the occurrence of concept drift by value of loss. When the new examples are more difficult to learn than previous ones or contain different properties or input modes from previous ones, it means the concept drift occurs and the loss of the new samples will substantially increase. At the same time, the model must update its knowledge and keep the maximum consistence with the knowledge learned before. When the loss is reduced, it indicates that the model can learn some meaningful features from the existing data streams and reach a stable state. However, in the process of network updating parameters and gradually reaching stability, it is possible that the loss exceeds the threshold and lead the network to be wrongly judged as the reappearance of concept drift. Therefore, we should not always judge whether the loss exceeds the threshold, but should estimate whether the concept drift occurs after the data reaches a stable state. As shown in Fig.4.

To detect concept drift, we use a sliding window to monitor the change of loss. When the mean value and variance of loss in sliding window are lower than a certain threshold value, we think that the model on the streaming data preserve a stationary state. At the same time, we record the mean and variance of the loss in the sliding window and the parameters of the network. However, one thing we need to remember is that, only after the stable state is reached, we can detect whether the concept drift appears. After the model reaches steady state, when the mean value and variance of loss in the sliding window are much higher than the ones of the stationary state, i.e.,$\mu(W)>\mu_L^{stable}+\sigma_L^{stable}$, we think that concept drift occurs in sliding window. 
\begin{figure}[!htbp]
	\centering
	\includegraphics[scale=0.7]{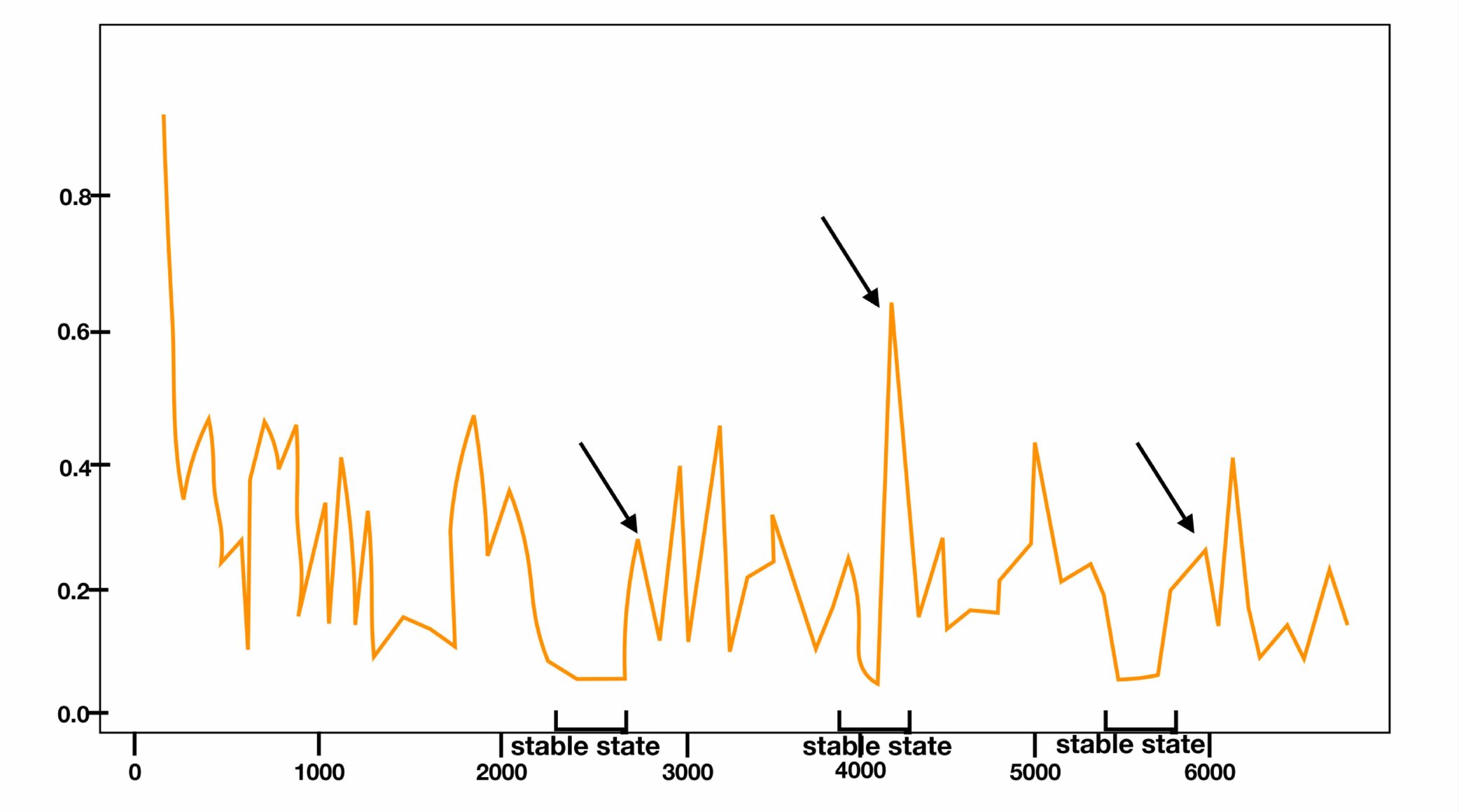}
	\caption{The x-axis represents input instances and y-axis represents the total loss. Arrows in the figure indicate that concept drift is detected. Only after the stable state is detected, can we detect whether the concept drift appears. Otherwise, it would lead to the miscalculation of the concept drift, that is, the adjustment state of the model would be judged as the concept drift. }
	\label{fig4}
\end{figure}

\subsubsection{Circumventing Catastrophic Forgetting}

Another problem called catastrophic forgetting describes the situation that the building model forgets the learned content on the previous data due to the retraining of the model on the new arriving streaming data. It requires the model on streaming data to make a trade-off between plasticity and stability. Usually, a natural solution to this problem is to incorporate a regularization term, such as L2 regularization term, in loss function to prevent the new learning model from deviating too much from the previous old one. However, the side effect of simply adding the L2 regularization term would prevent the model from learning the private features of the new approaching data, thus reducing the accuracy of the algorithm. In literature \cite{10kirkpatrick2017overcoming}, Kirkpatrick et al. proposed a method called elastic weight combination (EWC) to attack catastrophic forgetting. The reference object of this method is stochastic gradient descent (SGD) and L2 regularization with relative data parameters learned before. The core idea is to find out the importance of different parameters, which is represented by fisher information. The important parameters of data can be updated a little less, and the unimportant parameters can be updated a little more. This is relatively advanced for L2 regularities with the same constraints for all the parameters. However, if the network has a large number of parameters, it needs considerable systemic resources and induces a lot of computing work to record each parameter and evaluate its importance.

In this paper, we like killing two birds with one stone by dividing the features, which are built on outputs of the hidden layer, into shared features and private features. By doing this, we could benefit in two ways. On the one hand, we can extract better features. On the other hand, when the mode on streaming data is stable and invariant, we derive the shared features from outputs of the hidden layer, but when concept drift occurs, we utilize the previously memorized common features to prevent the network model from forgetting the previously useful information when learning the new streaming data. In addition, we also release the last three layers of the encoder for private feature learning, so that the model can improve the prediction accuracy of the algorithm and avoid catastrophic forgetting.

More specifically, when the data streams are stable, we save the parameters of the first three layers of the auto-encoder as the shared features. When concept drift occurs, we replace the first three layers of the auto-encoder with the previously saved shared features, and re-initialize the last three layers through the Xavier initialization method to extract the private features of the data stream.

\begin{algorithm}
     \renewcommand{\algorithmicensure}{\textbf{Output:}} 
	\caption{Adaptive dynamic update for network parameters.}
	\label{alg1}
	\begin{algorithmic}[1]
		\STATE  \textbf{Require:}streaming data, the thresholds of mean value and variance: $\delta_u$ and $\delta_\sigma$ , the set of network parameters: $\Omega$			
		\STATE  \textbf{Initialize:} set $\mu_L^{stable}=0$ , and $\sigma_L^{stable}=0,L^{sliding\_window}[]=\{\}$ , stable\_state=True.
		\STATE  \textbf{repeat}
		\STATE calculate the losses at $t,...,t+L-1$ imes using equation (11) in subsection 3.5, form the $L^{sliding\_window}[t,...,t+L-1]$.
		\STATE Calculate the mean $\mu_t(W)$ and variance $\sigma_t(W)$  of the losses at $t,...,t+L-1$ time.
		\STATE Detect whether there is the stable state:
		
		\textbf{if} stable\_state == True ,$\mu_t(W)<\delta_u$ and $\sigma_t(W)<\delta_\sigma$
		
		then let $\mu_L^{stable}=\mu_t(W)$ and $\sigma_L^{stable}=\sigma_t(W)$
		
		save the parameters of the first three auto-encoder layers to the set $\Omega$
		
		let stable\_state == False
		
		\textbf{end if}
		\STATE Calculate the mean $\mu_{t+1}(W)$ and variance $\sigma_t(W)$ of the losses at $t+1,...,t+L$ time
		\STATE Detect whether there is the concept drift:
		
		\textbf{if} stable\_state == False and $\mu_{t+1}(W)>\mu_L^{stable}+\sigma_L^{stable}$	:
		
		Replace the parameters of the first three auto-encoder layers with the previously saved one corresponding shared features layer in $\Omega$  
		
          Reinitialize the parameters of the last three auto-encoder layers through Xavier initialization method
          stable\_state == True
          
          \textbf{end if} 
		\STATE  \textbf{until} all training samples have been trained 	
		
		\ENSURE $\Omega$
	\end{algorithmic}

\end{algorithm}

In addition, the specific implementation procedure of adaptive dynamic update for network parameters is summarized in Algorithm I. In Algorithm I, we maintain an array. At current time $t$ ,array $L^{sliding\_window}[t,...,t+L-1]$ denotes the losses in current sliding window of size   ,whose elements are calculated by equation (11) in subsection 3.5; at next time $t+1$ ,array $L^{sliding\_window}[t+1,...,t+L]$ denotes the losses in next sliding window of size $L$ . At two consecutive instant, arrays $L^{sliding\_window}[t,...,t+L-1]$ and $L^{sliding\_window}[t+1,...,t+L]$ have overlaping $L-1$ elments. Meanwhile, at each time $t$ , we choose $S$  samples with the highest loss as the hard buffer and feed them into AOIL to retrain and update our model.

\section{The Specific Implementation of Proposed AOIL Algorithm}

\begin{figure}[!htbp]
	\centering
	\includegraphics[scale=0.7]{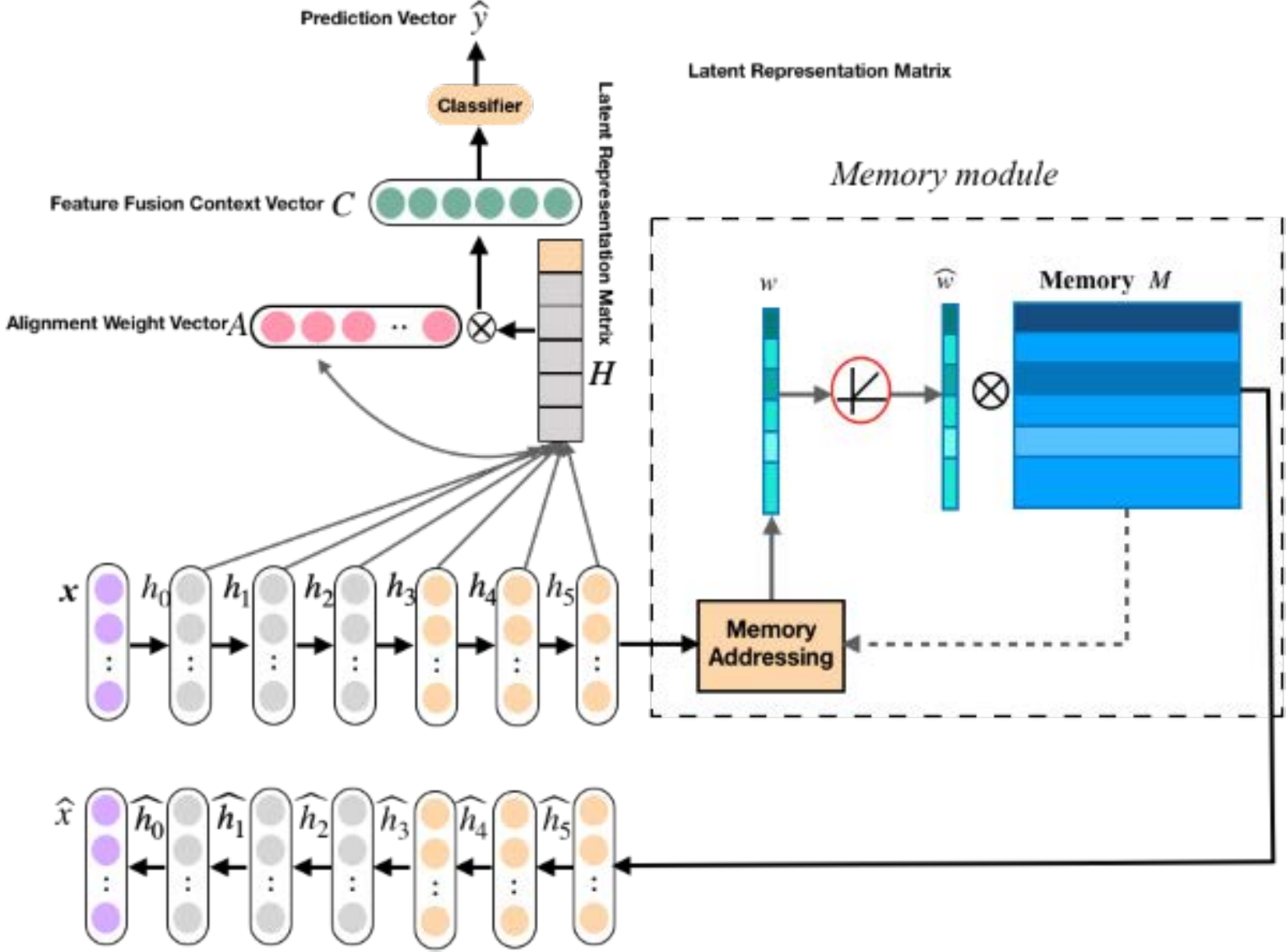}
	\caption{The illustration of the adaptive online continue learning for evolving data streams. We first get the latent representation of the input through the encoder, which is used to extract more useful information when building classifiers. To establish feature associations, we devise self-attention networks to enhance hidden layer correlations and fusion different hidden layer’s features. Then, by observing the change of the reconstruction loss of the auto-encoder, we could be aware of the change in the data distribution. Finally, according to our model adaptive algorithm, we dynamically adjust the network parameters.}
	\label{fig5}
\end{figure}

We are now at the position to summarize the results of previous discussions. As we can see in Fig.5, we propose a new adaptive online incremental learning structure, which introduces the concept of model adaptation. In light with the features derived from the hidden layers of encoder, the common features are shared by the data stream, while some private features are learned from a single data sample. In other words, features of the hidden layers are arranged in two kinds of attributes: shared attributes and private attributes. This mechanism enable AOIL effectively extract the characteristics of data, perform dynamic resource allocation which tracks the dynamic variation of data streams. Then, each hidden layer gets a fused feature through self-attention mechanism, connected to a softmax layer producing a global output. 

Note that the AOIL is developed in the online setting, data needs to be processed in a streaming fashion under the prequential test-then-train protocol. In addition, through back propagation and gradient descent, only one epoch traning on the streaming data is used to test and update the model parameters.

The specific implementation procedures of AOIL are shown in Algorithm II.

\begin{algorithm}
	\caption{Adaptive Online Incremental Learning for Evolving Data Streams.}
	\label{alg2}
	\begin{algorithmic}[1]
		\STATE  \textbf{Input:}streaming data: $\{x_i,y_i\}_{i=1}^N$		
		\STATE  \textbf{Repeat:} obtain the outputs $h_l,l=1,1,...,5$ of hidden layer according to the auto-encoder with short-cut connections\qquad\qquad\qquad\qquad\qquad\qquad\qquad\qquad\qquad\qquad\qquad\qquad\qquad\qquad  Eq.(1-2)
		\STATE Add the memory module $M$ to the auto-encoder\qquad\qquad\qquad\qquad\qquad\qquad\qquad\qquad  Eq.(3-6)
		\STATE Calculate the reconstruction loss with the memory module $L_{re}$\qquad\qquad\qquad\qquad\qquad Eq.(7)
		\STATE Concatenate the output of various hidden layer to form a latent representation matrix 
		
		$H=(h_0,h_1,...,h_L)$
		\STATE Obtain the alignment weight vector $A=[a_0,a_1,...,a_L]$\qquad\qquad\qquad\qquad\qquad\qquad\qquad Eq.(8)
		\STATE Obtain the feature fusion context vector $C$ \qquad\qquad\qquad\qquad\qquad\qquad\qquad\qquad  Eq.(9)
		\STATE Get the results of prediction value $\hat{y_t}$\qquad\qquad\qquad\qquad\qquad\qquad\qquad\qquad  Eq.(10)
		\STATE Calculate the total loss $L_{total}$\qquad\qquad\qquad\qquad\qquad\qquad\qquad\qquad\qquad\qquad Eq.(11)
		\STATE Dynamically update the network parameters via Algorithm I   
		\STATE \textbf{End}
		
	\end{algorithmic}

\end{algorithm}

\section{Experimental results}

A good adaptive online incremental learning algorithm on data streaming should meet the following requirements \cite{42pears2014detecting,43pesaranghader2016fast,44bifet2017classifier}: 1) Reduce misjudgment and false alarm in the case of the concept drift. High false positive rate requires more training data, which would consume more computing resources \cite{45vzliobaite2015towards}. Because the current estimated model does not reflect the changed distribution, the higher false negative rate reduces the classification accuracy. 2) It is necessary to detect concept drift quickly and reduce catastrophic forgetting and update its prediction model to maintain classification accuracy. 3) Robustness to noise – the algorithm must be able to distinguish concept drift from noise, and avoid confusing noise with concept drift and misjudgement, so as to achieve the desired detection effect.

It is worth mentioned that we have tried other approaches before reaching the current proposed one. In the initial design of the AOIL algorithm, we only adopt the auto-encoder and use the self attention mechanism to give different weights to each encoder layer to get the final output. Here, we call it as OIL-Base algorithm. In the following experiments, we also compare it with AOIL algorithm and find the performance of the OIL-Base algorithm is not ideal. Hence, we consequently consider adding adaptive dynamic update for network parameters (section 3.6). However, if the incoming streaming data exists concept drift which share common compositional patterns with the previous training streaming data or the decoder is ‘too strong’, which can decode the existing concept drift in data well, the encoder seems to have lost the function of detecting concept drift. In addition, through the experimental comparisons in subsection 5.6, we found that there is a significant difference the auto-encoder between with and without the memory unit. Motivated by this observation, we obtain the AOIL algorithm. Finally, we proposed AOIL-DAE algorithm to further improve the robustness of the AOIL algorithm.

In this section, we would carry out five experiments to validate the effectiveness of our algorithms. In the first experiment, we compared the advantages and disadvantages of different algorithms on seven datasets through five evaluation criteria. The five evaluation criteria are average accuracy, average precision, average recall, average F1, average AUC respectively. The comparative experimental results are obtained by running the open source code in the same computing environment. In order to make the experimental results more objective and fair, we have run each group of experiments ten times. In the second experiment, in order to verify that our algorithm effectively reduces the catastrophic forgetting problem, we divide the whole streaming data in five different stages :0-20\%,20-40\%,40-60\%,60-80\%, and 80-100\%. We investigate that as the number of learning samples increases,we compare the change of accuracy of different algorithms in last four stages. In the third experiment, to further illustrate the role of memory module in our proposed AOIL algorithm, we compared the accuracies of AOIL algorithm with or without memory module on different datasets under the same settings. In the fourth experiment, we discuss the convergence of our algorithm (AOIL). Finally, in the last experiment, we further improve the robustness of the algorithm by adding the denoising autoencoder to the original framework, and propose a new algorithm, named Adaptive Online Incremental Learning based on De-noising Auto-Encoder(AOIL-DAE). 

\subsection{Conditional Intensity Function}

In our algorithm, AOIL starts the learning process by having the 6-layer network, and set the number of neural unit of each hidden layer to 30. We use fully-connected layers as the encoder and decoder, expect the last one and memory moudle, Relu is used as the activation function for each hidden layer. The entire network parameters are updated using the Adam optimizer with a learning rate of 0.01\cite{46kingma2014method}. In the feature fusion part, the parameter $d_a$ for self-attention mechanism is a user-defined parameter, in our experiments, we set it to 30 to get good results. And in the memory module, the number of memory units $N$ can set to 50 that would lead to desirable results in all our experiments. Moreover, we set the length $L$ of the sliding loss window to 10, and we set $S$ equal to 5.

Due to the different properties among different datasets, we adopt different set-up for the threshold of mean value $\delta_u$ and variance $\delta_\sigma$ in AOIL algorithm. Usually, we can set $\delta_u\in (0.15,0.25)$ , and $\delta_\sigma =0.01$ to get good results. 

\subsection{Dataset Description}

The types of streaming data can be divided into stationary and non-stationary data, but most of them are non-stationary data in real environments, seven standard datasets with streaming data form are used in our experiment, we choose five groups of non-stationary data sets, two groups of stationary data sets for experimental comparison.

\textbf{Non-stationary datasets:}

(1) \emph{weather dataset}\cite{47ditzler2012incremental} : It is a dataset of weather forecast. The aim is to predict whether it will rain tomorrow, it belongs to the binary classification problem. The feature of the data are the temperature, pressure, visibility, wind speed, etc. This meteorological data set contains more than 50 years of weather change data. The concept drift for this data is due to seasonal and long-term climate change.

(2) \emph{sea dataset}\cite{48street2001streaming}: It is a dataset containing the abrupt and recurring drift. The initial data distribution of the first two features $f_1$ and $f_2$ are smaller than the threshold $Q$, e.g.,$f_1+f_2<Q$, threshold $Q$ changes in turn when the concept drift happens: $Q=4\rightarrow 7\rightarrow 4\rightarrow 7$.

(3) \emph{hyperplane dataset}\cite{7bifet2007learning}: Hyperplane is a synthetic streaming dataset generated by MOA software framework. It is a synthetic binary classification problem which has gradual drift, based on $d$ -dimensional random hyperplane $\sum_{j=1}^dw_jx_j>0$ . At first, data samples are extracted from some predefined probability distribution, and then gradually replaced by another probability distribution, so there is a transition period, the probability distribution obeyed by Hyperplane dataset exist the gradual drift.

(4) \emph{occupancy dataset}\cite{49candanedo2016accurate}: The occupancy dataset is about occupancy of rooms based on the environment of the room, this is a real-world dataset whose label is derived from time stamped pictures taken every minute, the dataset is also a non-stationary dataset. When the environment changes, the input distribution will change, which means there is covariate drift.

(5) \emph{kddcup dataset}\cite{50stolfo2000cost}: The data set is a binary classification problem to detect whether network connections encounter intrusion attacks. The non-stationary characteristics of the data set are due to the simulation of different types of network intrusion in the military network environment.

\textbf{Stationary datasets:}

(6) \emph{Hepmass dataset}\cite{51baldi2014searching}: This data set represents the physical experiments in Monte Carlo simulation to find the characteristics of strange particles with unknown mass. The classification task is to separate collision particles from the background source. A large number of data, about 10 million samples, have been generated in the study. Because they are stationary dataset, we select part of the data to do the experiment. In our experiment, only 200000 samples are used in our experiment and in doing so we envison that it does not affect on the experimental results.

(7) \emph{susy dataset} \cite{51baldi2014searching}: This data set represents the physical experiments in Monte Carlo simulation to distinguish between a signal process which produces supersymmetric particles and a background process which does not. The first eight features are kinematic property measured by particle detectors in accelerators. The last 10 features are the functions of the first 8 features; There are about 5 million samples in this data set, but because it is also a stationary data set, the probability distribution of the data will not change. To the experimental demonstration, we select 500000 samples for our experiment.

\subsection{Baseline Algorithms}

Considering that AOIL is an online incremental learning algorithm, several state-of-the-art methods are selected to compare, e.g., ALMA, aROMMA, aROW, CW, NAROW, NHERD, OGD, ROMMA, SCW,and SOP. In addition, some excellent methods focusing on the detection of concept drift are also in our comparative scope. e.g., Adwin, and PageHinkley. It should be noted that in our experiment, this two methods use Gaussian Bayesian approach as classifier. More specifically, the compared methods are: \textbf{ALMA}\cite{22gentile2001new}: Approximate Maximal Margin Classification Algorithm; \textbf{ROMMA} and \textbf{aROMMA}\cite{23li2002relaxed}: the Relaxed Online Maximum Margin Algorithm and its aggressive version; \textbf{aROW}\cite{24crammer2009adaptive}: the Adaptive Regularization of Weight Vectors; \textbf{CW}\cite{25crammer2008exact}: the Confidence-Weighted (CW) learning algorithm; \textbf{NAROW}\cite{26orabona2010new}: New variant of Adaptive Regularization; \textbf{NHERD} \cite{27crammer2010learning}: the Normal Herding method via Gaussian Herding; \textbf{OGD} \cite{28zinkevich2003online}: the Online Gradient Descent (OGD) algorithms; \textbf{SCW} \cite{29wang2012exact}: the recently proposed Soft Confidence Weighted algorithms; \textbf{SOP} \cite{30cesa2005second}: second-order perception; \textbf{Gaussian Bayesian}: the classifer in the Adwin, and PageHinkley algorithm; \textbf{Adwin} \cite{5ren2017life}: A concept drift detecting method, using adaptive online sliding windows according to the rate of change observed from the data; \textbf{PageHinkley} \cite{31gama2013evaluating}: Another concept drift detecting method, learning from time varying data with adaptive window; \textbf{OIL-Base}: a model trained as the AOIL but without the memeory moudle and the adaptive mechanism.

\subsection{Performance Evaluation Criteria on Different Datasets}

In this section, five criteria are taken to evaluate the performance of our algorithm. We take the average of these values for comparison, the comparision results are reported in Table 1. From the Table 1, it can be obviously found that our AOIL algorithm achieves better results on different datasets using different evaluation criteria. In addition, it should be noted that the Adwin and PageHinkley algorithm, use the Gaussian Bayesian (GaussianNB) as the classifer, can not play a full role when these two algorithms do not detect the concept drift effectively, or when dealing with stable datasets. Through the analysis of the experimental results, it reveals that the AOIL is a high-competitive approach to handle both the stationary data and non-stationary data.

\begin{table}[!htbp]
	\centering
	\caption{The Comparative Results of Different Algorithms on Different Datasets.}
	\label{tb1}
	\begin{tabular}{cccccccc}\hline
		\multirow{2}{*}{\textbf{method}} &\multicolumn{7}{c}{Average Accuracy}\\
		\cline{2-8}
		 &weather  &sea  &hyperplane  &occupancy  &kddcup  &hepmass  &susy\\\hline
		AOIL	&77.44\%	&91.71\%	&91.03\%	&98.15\%	&99.74\%	&83.28\%	&79.37\%\\
          ALMA	&76.33\%	&76.73\%	&66.60\%	&96.98\%	&99.49\%	&83.16\%	&79.18\%\\
          aROMMA	&74.89\%	&33.79\%	&43.99\%	&96.65\%	&99.43\%	&81.99\% 	&76.20\%\\
          aROW	&73.96\%	&76.81\%	&66.42\%	&96.94\%	&99.43\%	&82.70\%	&77.57\%\\
          CW	&69.26\%	&43.84\%	&43.34\%	&96.74\%	&99.29\%	&50.45\%	&56.25\%\\
          NAROW	&54.32\%	&68.62\%	&64.88\%	&89.24\%	&99.42\%	&79.98\%	&68.34\%\\
          NHERD	&73.69\%	&76.72\%	&66.42\%	&96.75\%	&99.39\%	&82.65\%	&77.31\%\\
          OGD	&71.15\%	&76.85\%	&66.51\%	&96.26\%	&99.29\%	&82.52\%	&78.08\%\\  
          ROMMA	&73.85\%	&33.81\%	&43.99\%	&96.54\%	&99.35\%	&81.74\%	&76.21\%\\
          SCW	&72.19\%	&76.71\%	&66.12\%	&96.86\%	&99.44\%	&82.65\%	&77.78\%\\
          SOP	&69.88\%	&77.95\%	&70.34\%	&96.37\%	&98.89\%	&78.72\%	&76.81\%\\
          GassianNB	&69.19\%	&88.41\%	&84.97\%	&97.90\%	&97.28\%	&80.09\%	&73.56\%\\
          Adwin	&70.58\%	&91.35\%	&89.76\%	&92.56\%	&97.28\%	&80.09\%	&73.56\%\\
          PageHigkley     &69.28\%	&88.16\%	&84.97\%	&97.90\%	&97.28\%	&80.09\%	&73.56\%\\
          OIL-Base	&71.83\%	&87.11\%	&88.07\%	&82.93\%	&99.01\%	&82.00\%	&78.56\%\\\hline
	\end{tabular}
\end{table}
\begin{table}[!htbp]
	\centering
	\begin{tabular}{cccccccc}\hline
		\multirow{2}{*}{\textbf{method}} &\multicolumn{7}{c}{Average Precision}\\
		\cline{2-8}
		 &weather  &sea  &hyperplane  &occupancy  &kddcup  &hepmass  &susy\\\hline
		AOIL	&0.7992	&0.9001	&0.8750	&0.9901	&0.9980	&0.7800	&0.7464\\
ALMA	&0.5158	&0.7909	&0.6700	&0.8479	&0.9968	&0.7760	&0.7170\\
aROMMA	     &0.5128	&0.7307	&0.6065	&0.8688	&0.9955	&0.7610	&0.6685\\
aROW	&0.4801	&0.7873	&0.6672	&0.8774	&0.9968	&0.7691	&0.6990\\
CW	&0.4389	&0.7152	&0.6014	&0.8717	&0.9946	&0.4993	&0.4843\\
NAROW	&0.3986	&0.7099	&0.6373	&0.6311	&0.9966	&0.7371	&0.5948\\
NHERD	&0.4810	&0.7825	&0.6671	&0.8701	&0.9969	&0.7683	&0.6984\\
OGD	&0.4108	&0.7886	&0.6691	&0.8540	&0.9978	&0.7703	&0.7061\\
ROMMA	&0.4990	&0.7309	&0.6065	&0.8648	&0.9947	&0.7580	&0.6685\\
SCW	&0.4567	&0.7923	&0.6683	&0.8747	&0.9967	&0.7723	&0.6954\\
SOP	&0.4448	&0.8278	&0.7105	&0.8577	&0.9921	&0.7247	&0.6768\\
GassianNB	     &0.4273	&0.8651	&0.7972	&0.9138	&0.9723	&0.7307	&0.6536\\
Adwin	 &0.4526	&0.8961	&0.8573	&0.7210	&0.9723	&0.7307	&0.6536\\
PageHigkley	&0.4416	&0.8591	&0.7972	&0.9138	&0.9723	&0.7307	&0.6536\\
OIL-Base	&0.7205	     &0.7513	&0.8394	&0.8391	&0.9571	&0.7552	&0.7377\\\hline
	\end{tabular}
\end{table}
\begin{table}[!htbp]
	\centering
	\begin{tabular}{cccccccc}\hline
		\multirow{2}{*}{\textbf{method}} &\multicolumn{7}{c}{Average Recall}\\
		\cline{2-8}
		 &weather  &sea  &hyperplane  &occupancy  &kddcup  &hepmass  &susy\\\hline
AOIL	&0.8882	&0.9685	&0.9032	&0.9827	&0.9973	&0.8675	&0.8731\\
ALMA	&0.5107	&0.9008	&0.7850	&0.9408	&0.9960	&0.8244	&0.6865\\
aROMMA	     &0.6362	&0.0820	&0.067	&0.9175	&0.9966	&0.8195	&0.7653\\
aROW	&0.4779	&0.9117	&0.7933	&0.9453	&0.9951	&0.8270	&0.6531\\
CW	&0.5272	&0.3555	&0.0607	&0.9175	&0.9956	&0.0129	&0.0859\\
NAROW	&0.8614	&0.9493	&0.9216	&0.8282	&0.9951	&0.8063	&0.5154\\
NHERD	&0.5085	&0.9221	&0.7942	&0.9510	&0.9942	&0.8262	&0.6346\\
OGD	&0.1731	&0.9092	&0.7862	&0.9091	&0.9916	&0.8095	&0.6555\\
ROMMA	&0.6217	&0.0825	&0.0676	&0.9072	&0.9965	&0.8176	&0.7645\\
SCW	&0.4594	&0.8966	&0.7721	&0.9442	&0.9955	&0.8090	&0.6951\\
SOP	&0.5290	&0.8508	&0.7447	&0.9101	&0.9926	&0.7865	&0.7500\\
GassianNB	     &0.5842	&0.9478	&0.8507	&0.9591	&0.9330	&0.8654	&0.5564\\
Adwin	&0.7058	&0.9619	&0.8974	&0.6835	&0.9930	&0.8654	&0.5564\\
PageHigkley	&0.6570	&0.9548	&0.8507	&0.9591	&0.9930	&0.8654	&0.5564\\
OIL-Base	&0.7951	     &0.7708	&0.8807	&0.9656	&0.9692	&0.8575	&0.8641\\
\hline
	\end{tabular}
\end{table}
\begin{table}[!htbp]
	\centering
	\begin{tabular}{cccccccc}\hline
		\multirow{2}{*}{\textbf{method}} &\multicolumn{7}{c}{Average F1}\\
		\cline{2-8}
		 &weather  &sea  &hyperplane  &occupancy  &kddcup  &hepmass  &susy\\\hline
AOIL	&0.8477	&0.9395	&0.9031	&0.9882	&0.9972	&0.8384	&0.8305\\
ALMA	&0.5881	&0.8470	&0.7354	&0.9284	&0.9968	&0.8298	&0.7529\\
aROMMA	     &0.6264	&0.1508	&0.1250	&0.9217	&0.9964	&0.8192	&0.7480\\
aROW	&0.5484	&0.8490	&0.7364	&0.9303	&0.9964	&0.8264	&0.7290\\
CW	&0.5315	&0.4751	&0.1125	&0.9240	&0.9956	&0.0254	&0.1536\\
NAROW	&0.5551	&0.8122	&0.7563	&0.7687	&0.9963	&0.8004	&0.6006\\
NHERD	&0.5612	&0.8499	&0.7366	&0.9268	&0.9961	&0.8259	&0.7210\\
OGD	&0.2842	&0.8488	&0.7351	&0.9124	&0.9955	&0.8217	&0.7342\\
ROMMA	&0.6113	&0.1513	&0.1250	&0.9190	&0.9959	&0.8168	&0.7480\\
SCW	&0.5223	&0.8462	&0.7293	&0.9286	&0.9965	&0.8228	&0.7429\\
SOP	&0.5375	&0.8450	&0.7481	&0.9155	&0.9931	&0.7864	&0.7492\\
GassianNB	     &0.5435	&0.9116	&0.8500	&0.9513	&0.9832	&0.8130	&0.6591\\
Adwin       	&0.6018	&0.9334	&0.8977	&0.7971	&0.9832	&0.8130	&0.6591\\
PageHigkley	&0.5732	&0.9104	&0.8500	&0.9513	&0.9832	&0.8130	&0.6591\\
OIL-Base	     &0.8095	&0.8152	&0.8786	&0.8990	&0.9752	&0.8264	&0.8135\\
\hline
	\end{tabular}
\end{table}
\begin{table}[!htbp]
	\centering
	\begin{tabular}{cccccccc}\hline
		\multirow{2}{*}{\textbf{method}} &\multicolumn{7}{c}{Average AUC}\\
		\cline{2-8}
		 &weather  &sea  &hyperplane  &occupancy  &kddcup  &hepmass  &susy\\\hline
AOIL	&0.7321	&0.8973	&0.9114	&0.9798	&0.9952	&0.8329	&0.7945\\
ALMA	&0.6994	&0.6667	&0.6394	&0.9585	&0.9933	&0.8316	&0.7844\\
aROMMA	     &0.7204	&0.5306	&0.5230	&0.9465	&0.9907	&0.8199	&0.7622\\
aROW	&0.6734	&0.6599	&0.6353	&0.9607	&0.9932	&0.8270	&0.7671\\
CW	&0.6507	&0.5008	&0.5165	&0.9493	&0.9888	&0.5024	&0.5288\\
NAROW	&0.6237	&0.4878	&0.5879	&0.8691	&0.9928	&0.8599	&0.6715\\
NHERD	&0.6792	&0.6504	&0.6351	&0.9616	&0.9933	&0.8265	&0.7633\\
OGD	&0.5754	&0.6625	&0.6380	&0.9403	&0.9948	&0.8251	&0.7719\\
ROMMA	&0.7090	&0.5308	&0.5229	&0.9444	&0.9808	&0.8174	&0.7622\\
SCW	&0.6555	&0.6695	&0.6365	&0.9598	&0.9929	&0.8264	&0.7719\\
SOP	&0.6559	&0.7332	&0.6942	&0.9443	&0.9834	&0.7872	&0.7668\\
GassianNB	     & 0.6627	&0.8616	&0.8497	&0.9717	&0.9424	&0.8008	&0.7221\\
Adwin	0.7065	&0.8964	&0.8976	&0.8375	&0.9424	&0.8008	&0.7221\\
PageHigkley	&0.6831	&0.8557	&0.8497	&0.9717	&0.9424	&0.8008	&0.7221\\
OIL-Base	     &0.5958	&0.8503	&0.8807	&0.6458	&0.9823	&0.8200	&0.7785\\
\hline
	\end{tabular}
\end{table}

\subsection{The Incremental Quantities of Accuracies at Different Learning Stages for entire streaming dataset}

In this subsection, we compare the incremental amounts of accuracies of different algorithms that perform on the whole streaming data of 20-40\%,40-60\%,60-80\% , and 80-100\%. As the example size increases, we plot the experimental results in heat maps, in which the x-axis represents the example size in percentage, and the y-axis represents different algorithms. The different color represents different incremental quantities of accuracies. As we can see in the Fig.6, on the whole, our algorithm is in a state of positive growth in all stages, while, for some dataets, such as weather data, the ROMMA algorithm has a negative incremental quantities of accuracies, which represents the data streams may appear concept drift, but the model does not deal with it well, thus lead to a negative increase in accuracies. However, our AOIL algorithm can effectively detect and adaptively adjust the network parameters, and ultimately achieve good results. 
\begin{figure}[!htbp]
	\centering
	\includegraphics[scale=0.7]{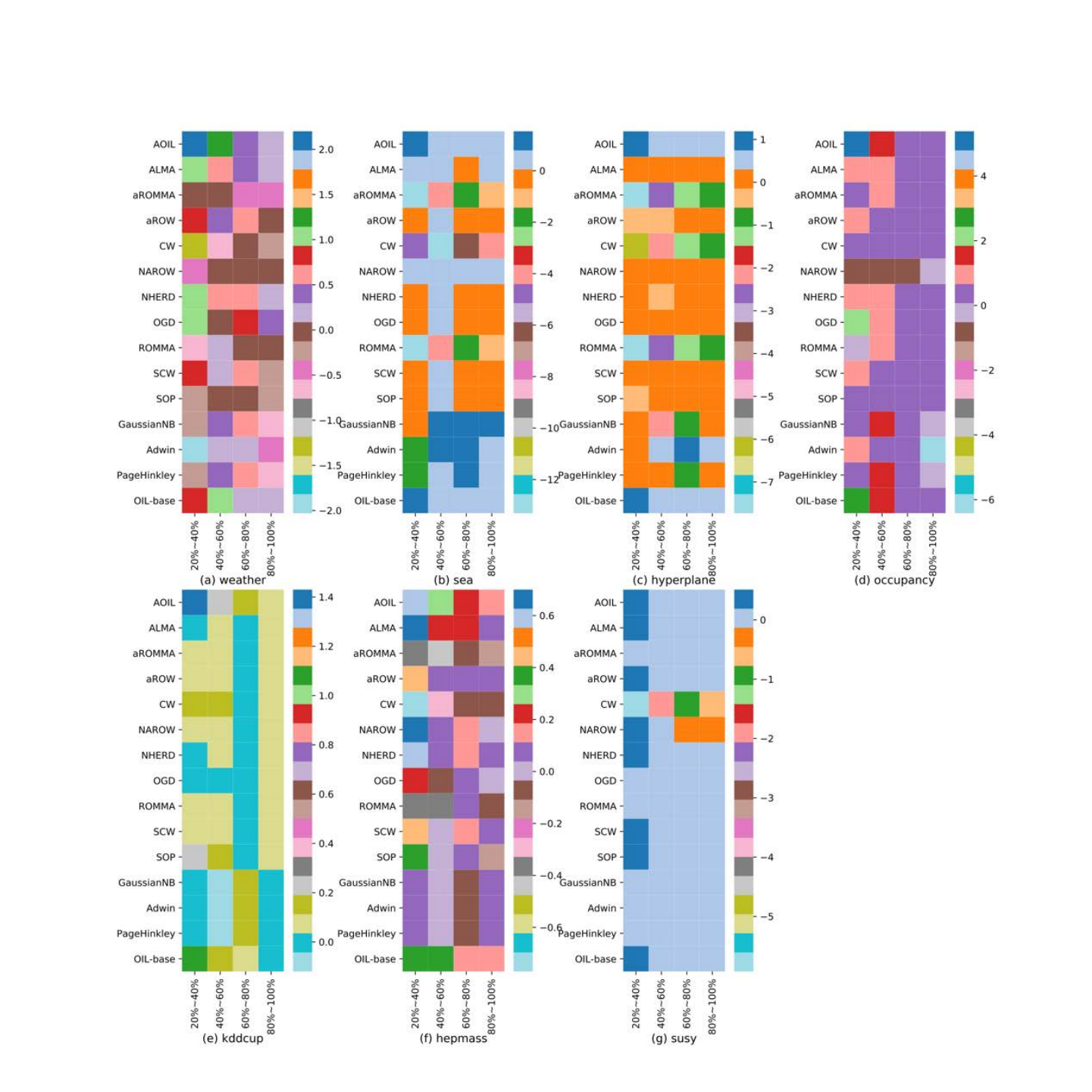}
	\caption{The heat-map of the incremental quantities of accuracies for four algorithms that perform on entire streaming datasets of 20-40\%, 40-60\%, 60-80\% and 80-100\%. The x-axis represents the sample size in percentage, and the y-axis represents different algorithms.}
	\label{fig6}
\end{figure}

\subsection{ The AOIL Algorithm With or Without Memory Module}

In this subsection, in order to show that it is necessary and effective to add the memory module to the auto-encoder in our algorithm (AOIL), we conduct experiments on different data sets, and compare the algorithm with memory module (AOIL) and the algorithm without memory module (AOIL\_NO\_MEMORY). From the comparison results in the Fig.7, we find that after adding memory module, the accuracies of AOIL algorithm is higher than the AOIL\_NO\_MEMORY algorithm. To a certain extent, it shows that when the data probability distribution changes, we can better detect the emergence of concept drift by using the reconstruction loss of autoencoder with memory moudule, so as to trigger the update mechanism and improve the predictive accuracies of the algorithm.

\begin{figure}[!htbp]
	\centering
	\subfigure[weather]{
		\includegraphics[width=7cm]{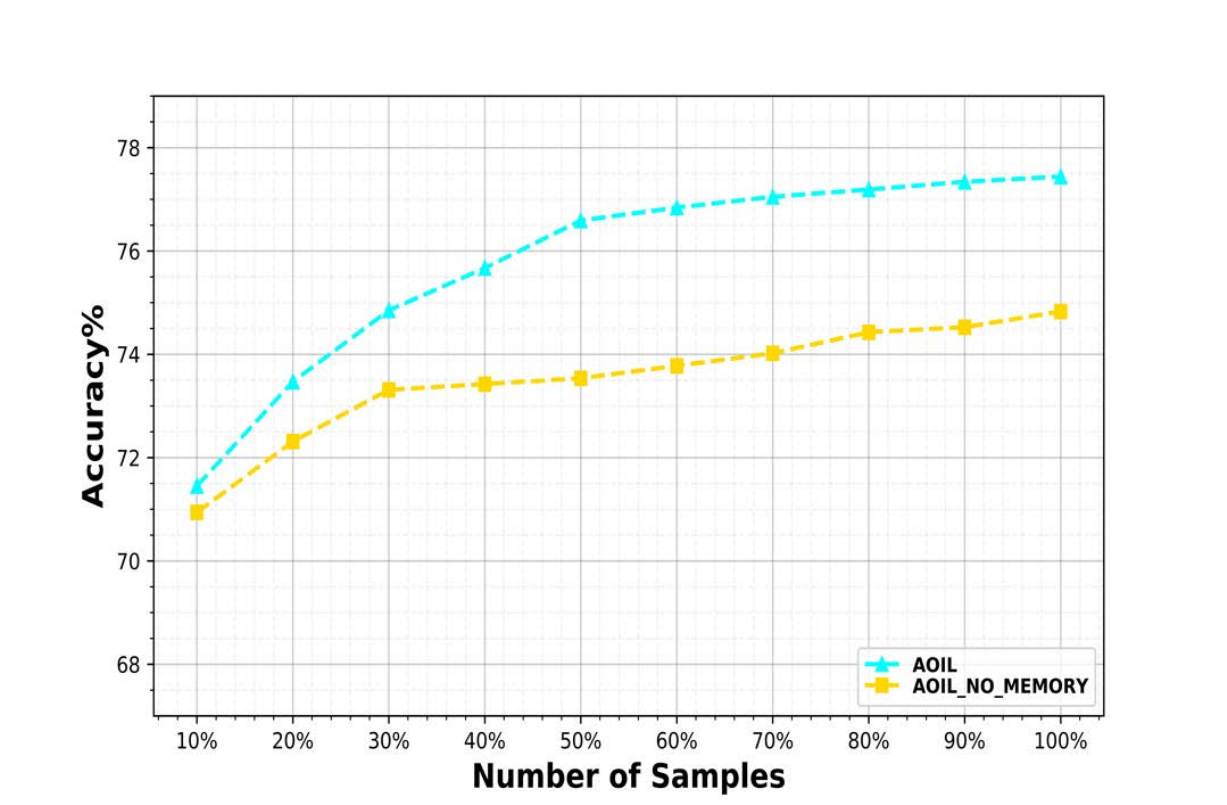}
	}
	\quad
	\subfigure[sea]{
		\includegraphics[width=7cm]{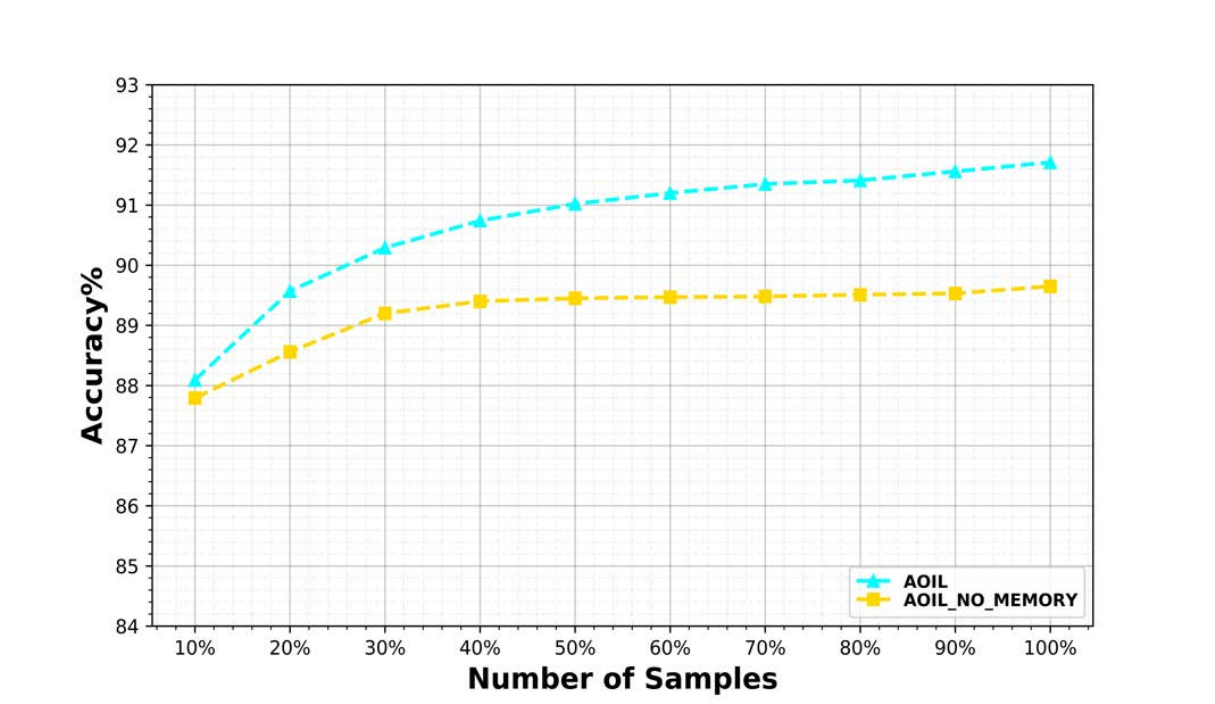}
	}
	\quad
	\subfigure[hyperplane]{
		\includegraphics[width=7cm]{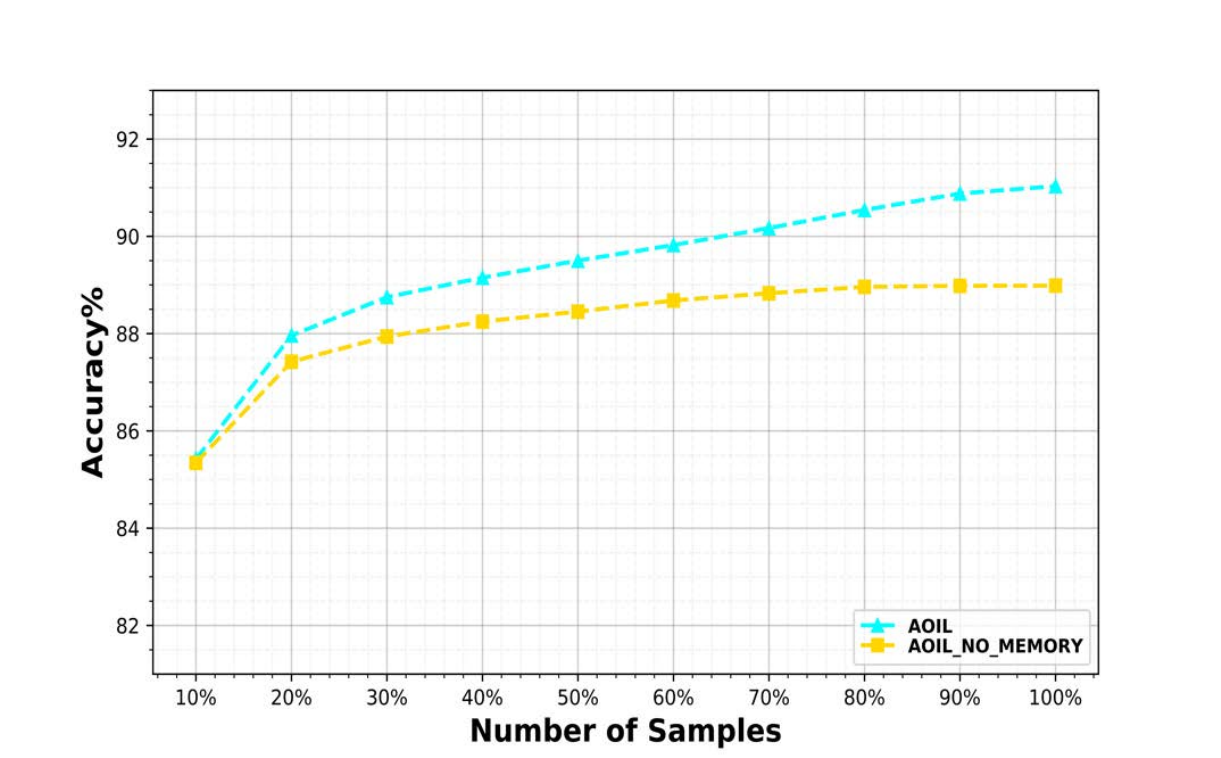}
	}
	\subfigure[occupancy]{
		\includegraphics[width=7cm]{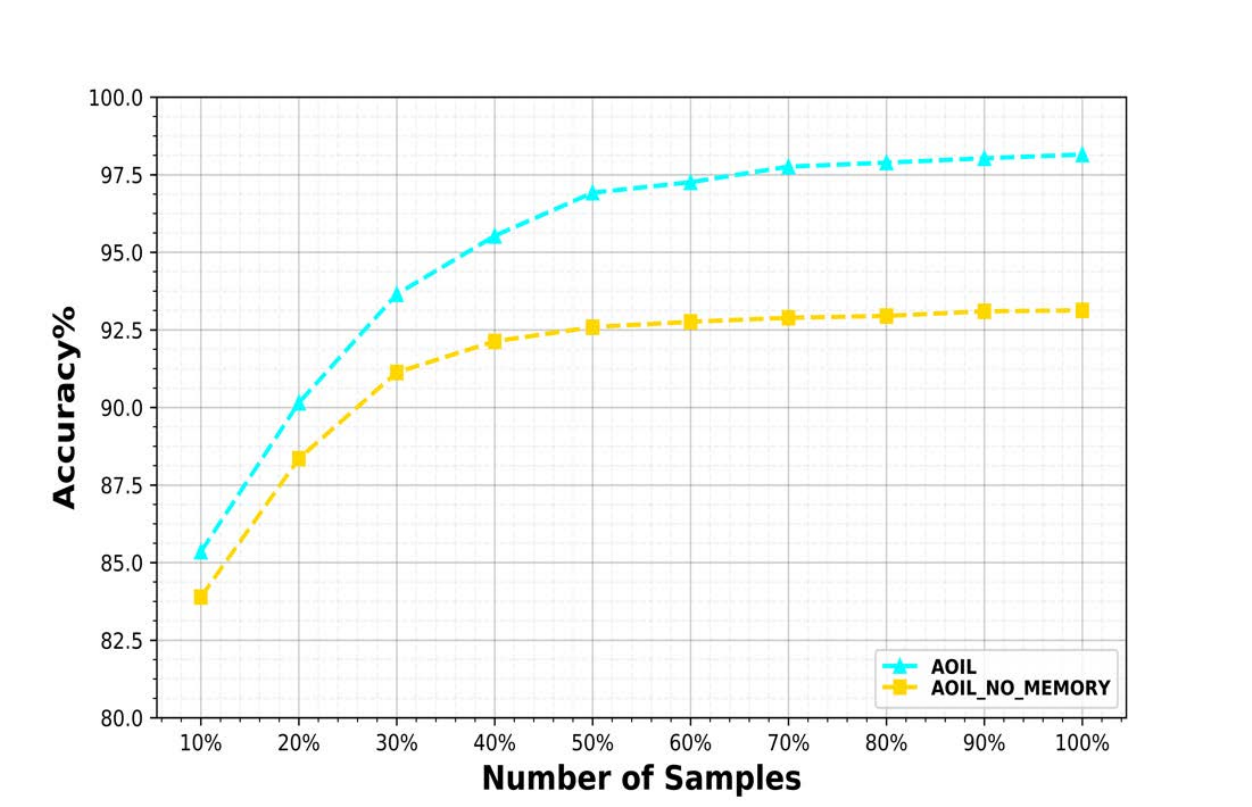}
	}
	\quad
	\subfigure[kddcup]{
		\includegraphics[width=7cm]{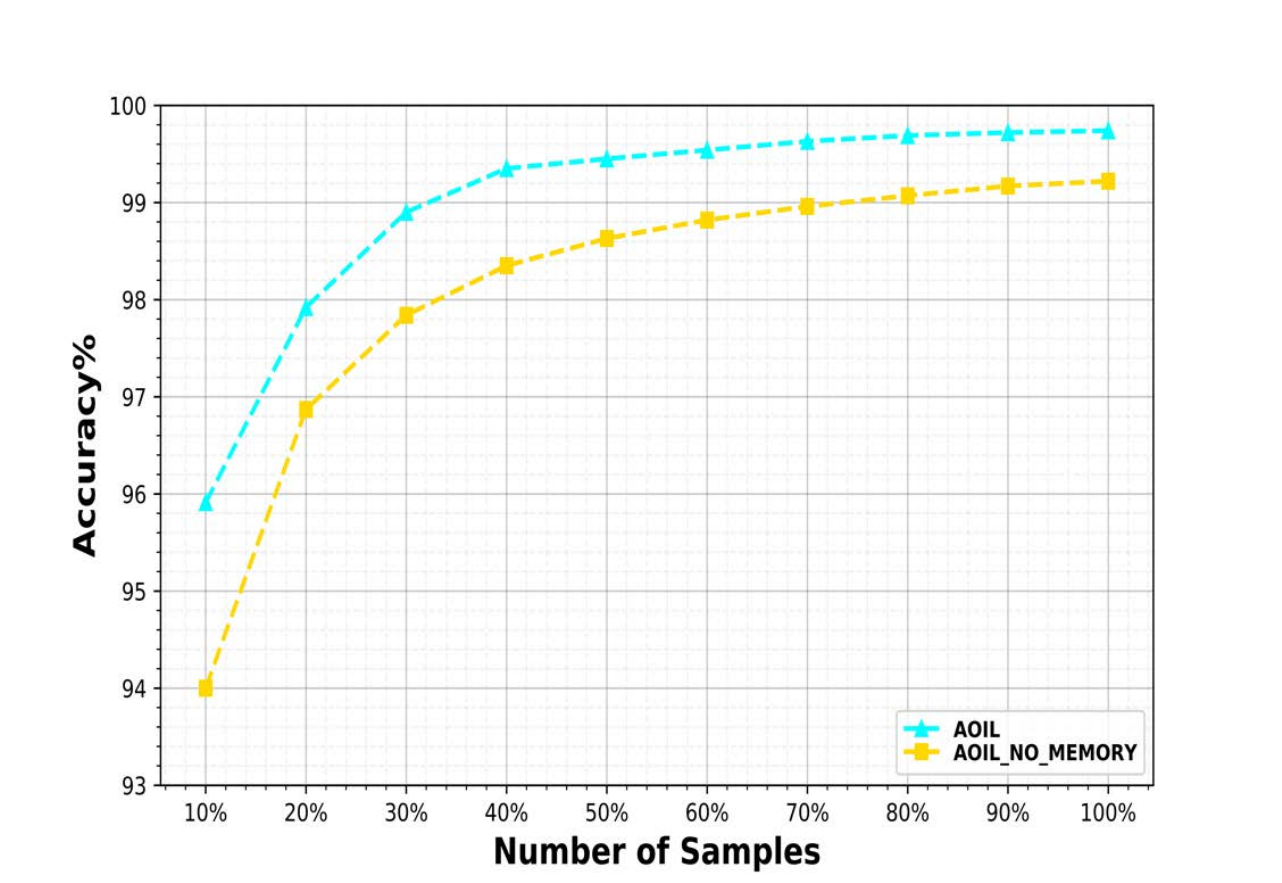}
	}
	\quad
	\subfigure[hepmass]{
		\includegraphics[width=7cm]{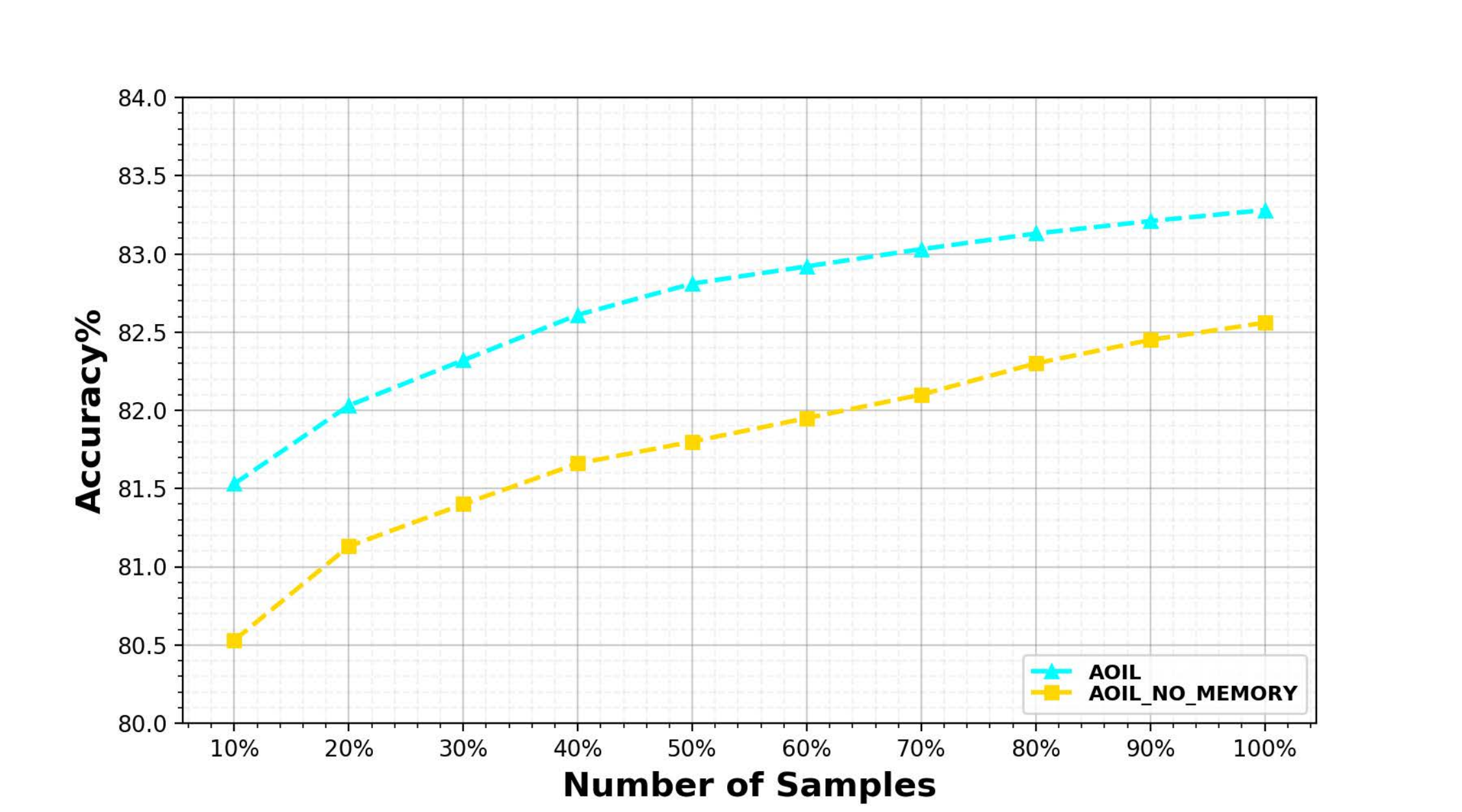}
	}
	\quad
	\subfigure[susy]{
	\includegraphics[width=7cm]{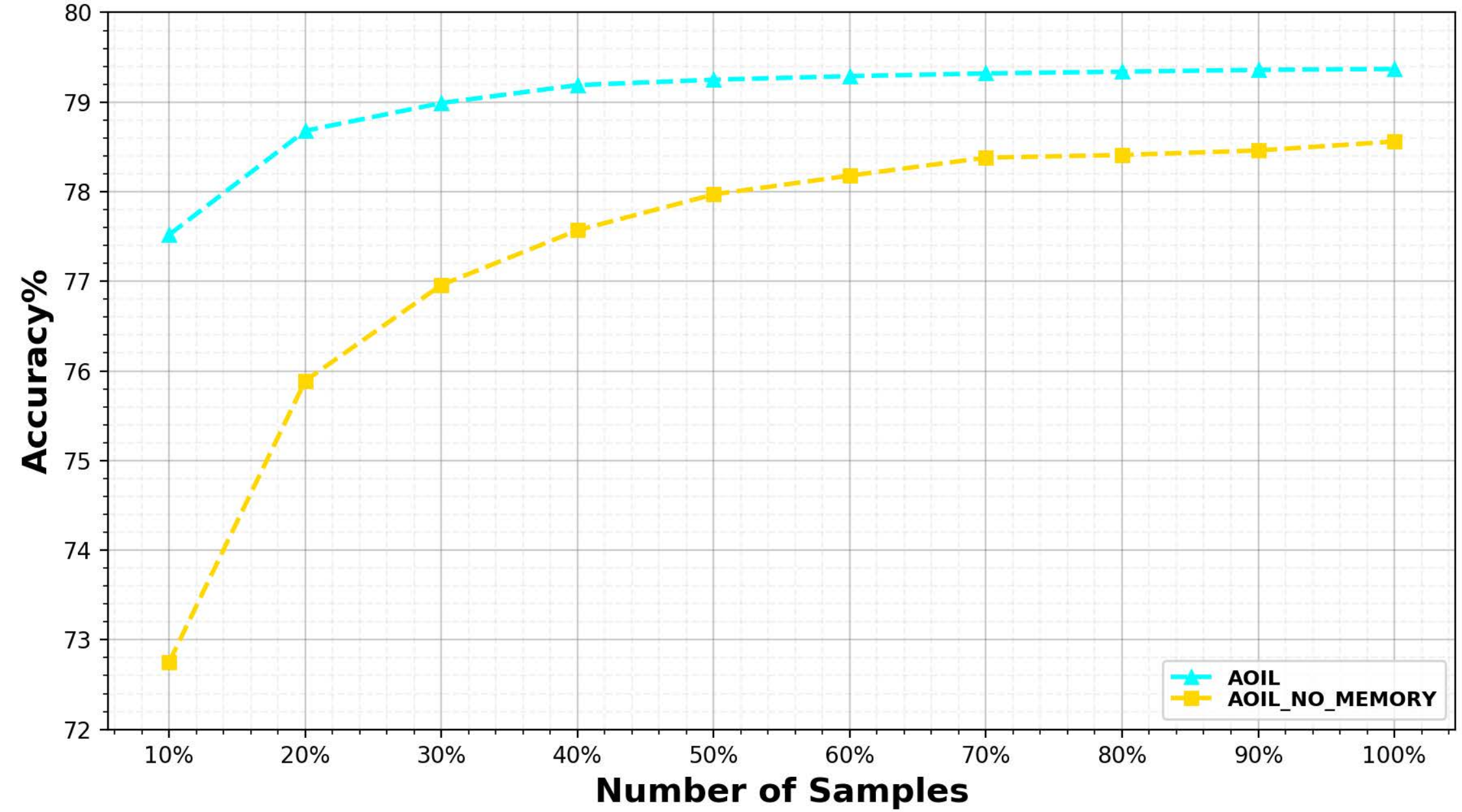}

	}
	\caption{Compare the accuracy of AOIL algorithm with or without memory module on different datasets. }
\end{figure}

\subsection{Convergence analysis of the algorithm }

In this subsection we show the convergence of the AOIL algorithm. Due to the characteristics of online incremental learning, we only train one epoch for all datasets. As we can see in Fig.8, the loss curve $L_{total}$ may fluctuate due to the concept drift, but the trend of total loss would converge to a certain range.
\begin{figure}[!htbp]
	\centering
	\includegraphics[scale=0.7]{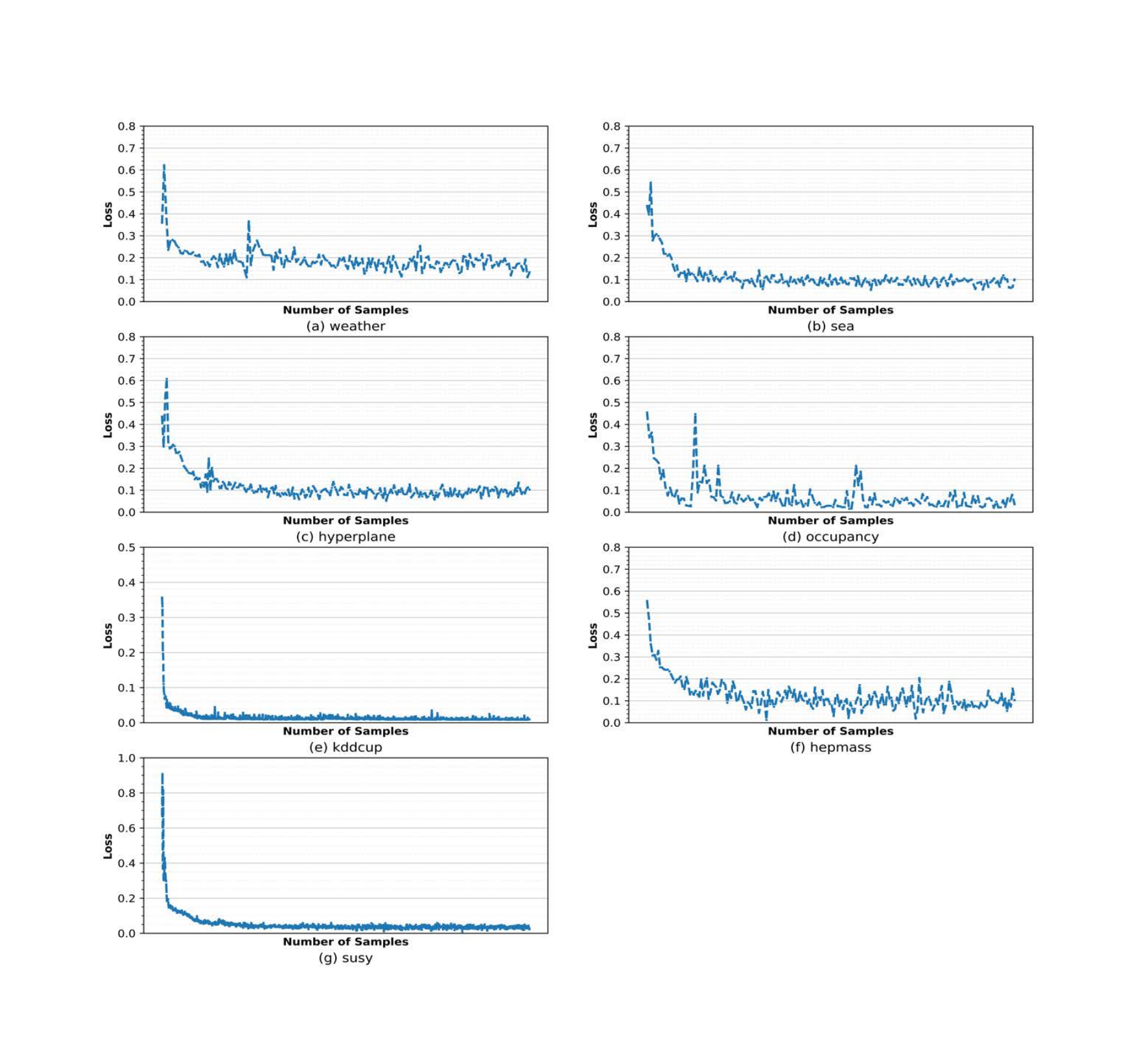}
	\caption{The convergence of the AOIL algorithm on different datasets. }
	\label{fig8}
\end{figure}

\subsection{Robustness analysis of the algorithm}

In fact, in the real scenario, the inputs are often corrupted by noise, and the presence of the noise often requires the model to have certain robustness. To further illustrate the anti-noise ability of our algorithm, we conduct the following experiments. Random noise is added to 20\% of the raw data. More specifically, we inject masking noise with Gaussian distribution with a mean of 0 and a variance of 0.1 to the data. In order to improve the processing ability of noisy data, we propose a new algorithm, named Adaptive Online Incremental Learning based on De-noising Auto-Encoder(AOIL-DAE), adding the de-nosing auto-encoder to the original AOIL network structure. This is done by first corrupting the initial input $x$ into $\tilde{x}$ in the form of a stochastic mapping $\tilde{x}\sim q_D(\tilde{x}\mid x)$ . The collapsed input $\tilde{x}$ is then mapped, as the same process as auto-encoder does before, to a hidden representation. In addition, the following process of AOIL-DAE is the same as AOIL algorithm. 

We compare the performance of the AOIL and AOIL-DAE on noisy data. The experimental results can be seen in Table.2 and Fig.8. From Table.2, we can see that our algorithm performs well compared to other state-of-art algorithms. Moreover, in order to make a more obvious comparison, we tend to show the predictive performance changes, i.e. the increase and decrease of each performance evaluation criteria after adding the noise, through the radar map, as shown in Fig.9. 
\begin{figure}[!htbp]
	\centering
	\subfigure[weather(noise)]{
		\includegraphics[width=4.5cm]{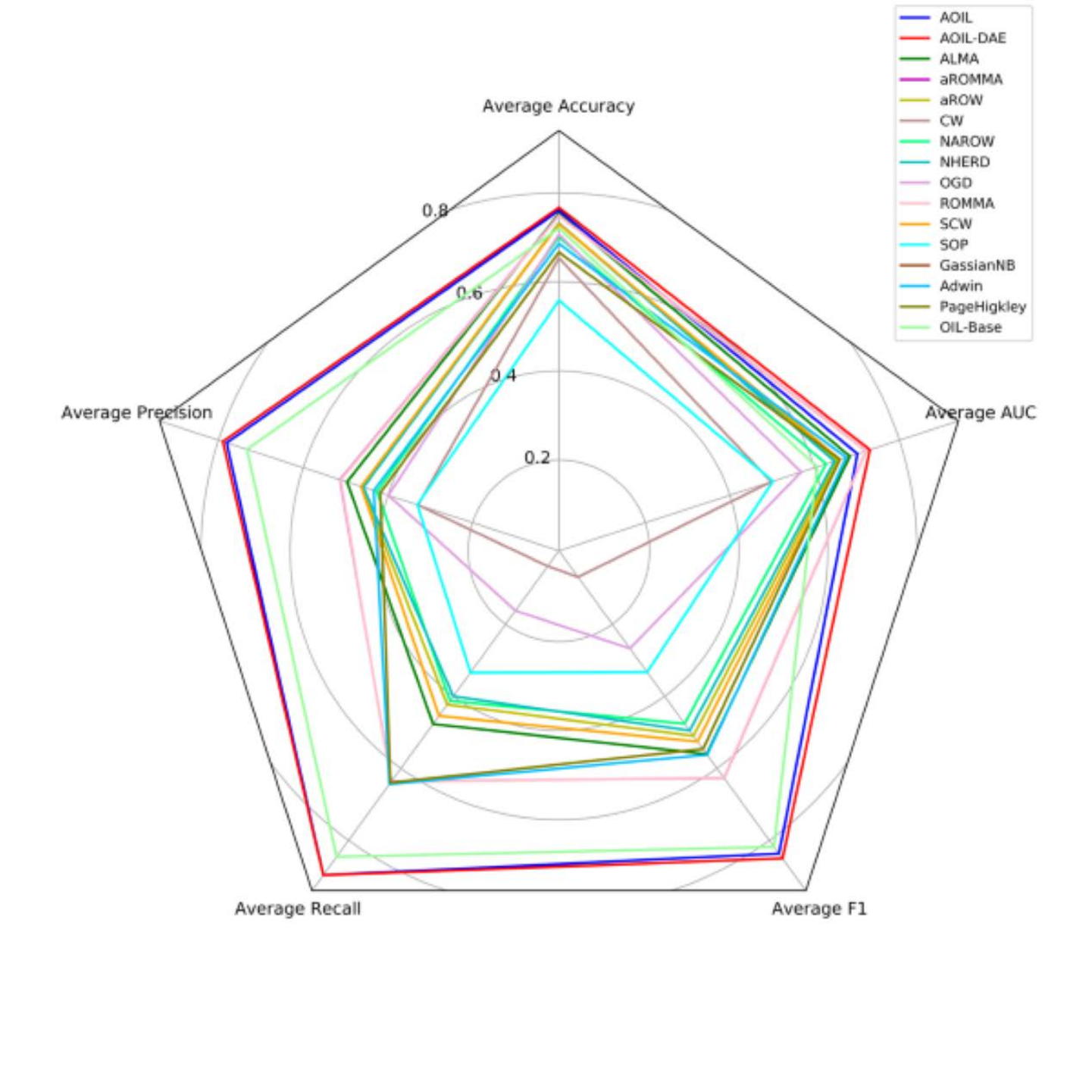}
	}
	\quad
	\subfigure[sea(noise)]{
		\includegraphics[width=4.5cm]{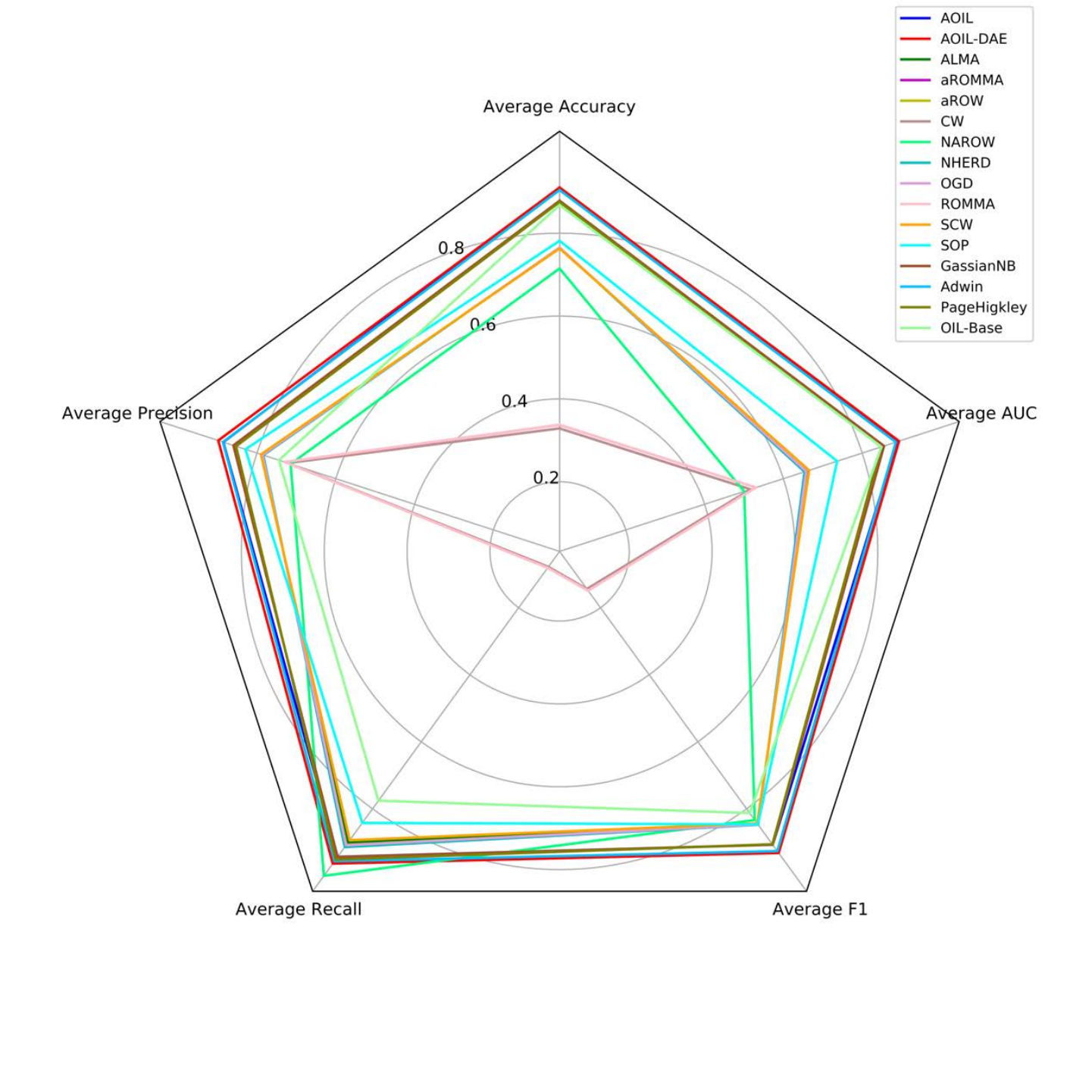}
	}
	\quad
	\subfigure[hyperplane(noise)]{
		\includegraphics[width=4.5cm]{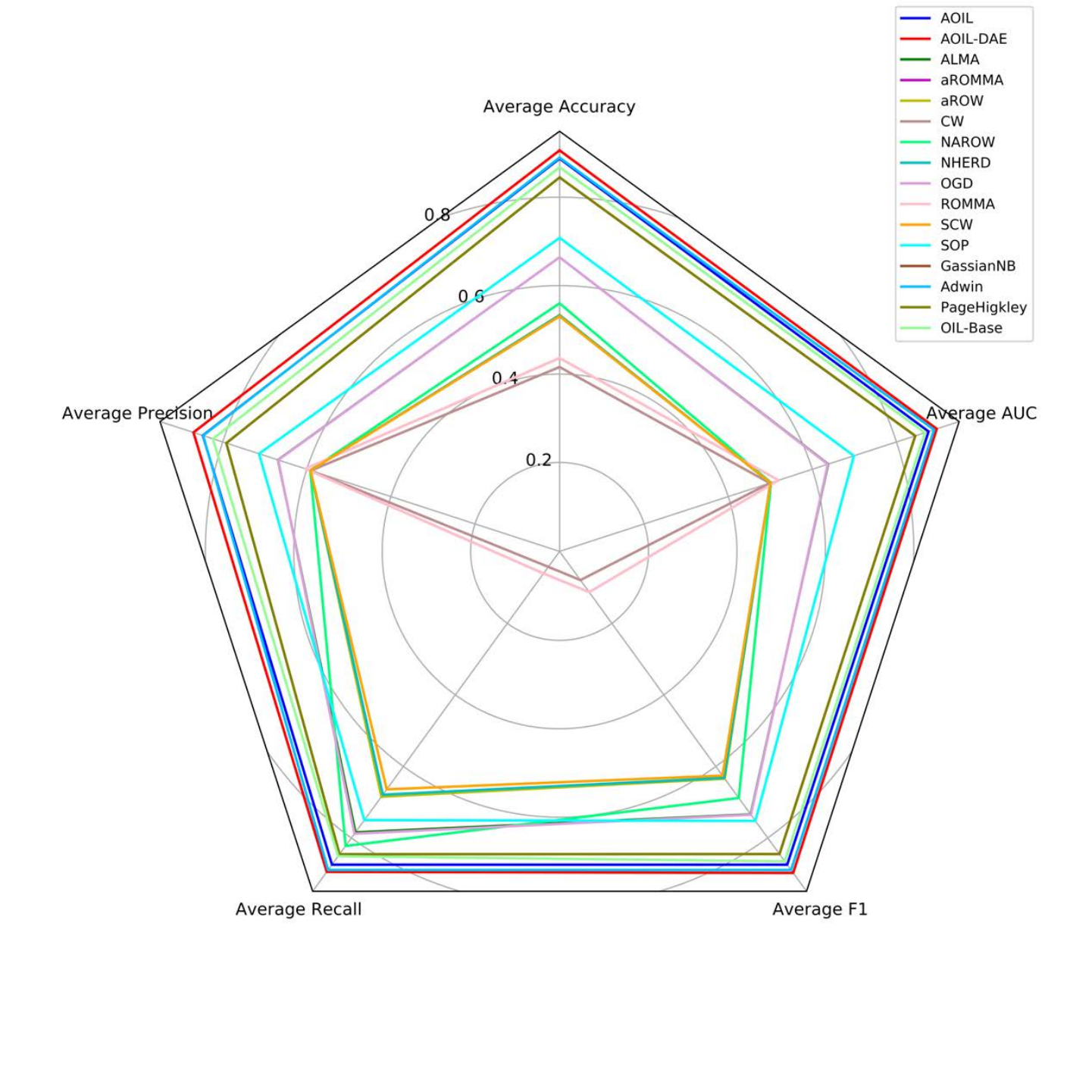}
	}
	\subfigure[occupancy(noise)]{
		\includegraphics[width=4.5cm]{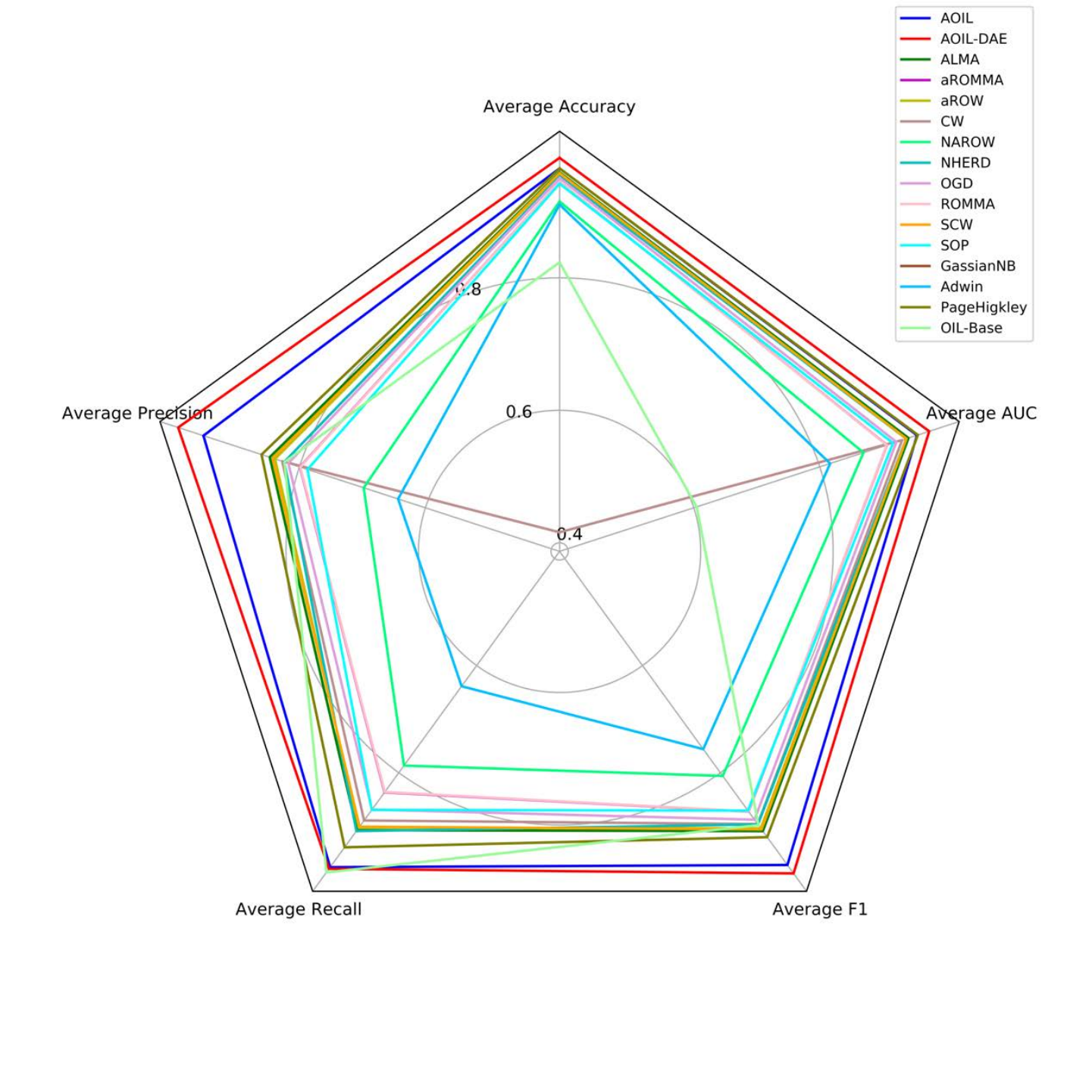}
	}
	\quad
	\subfigure[kddcup(noise)]{
		\includegraphics[width=4.5cm]{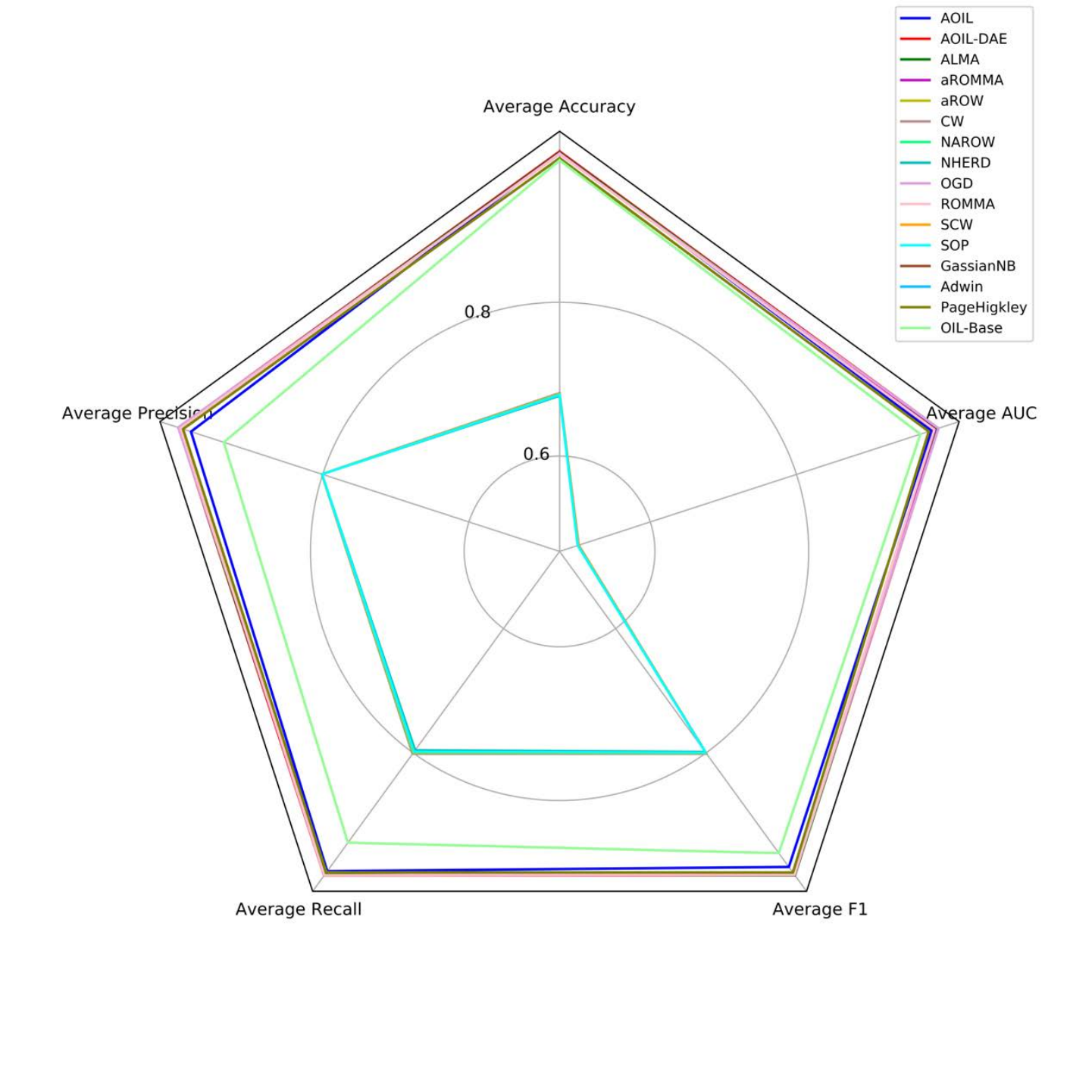}
	}
	\quad
	\subfigure[hepmass(noise)]{
		\includegraphics[width=4.5cm]{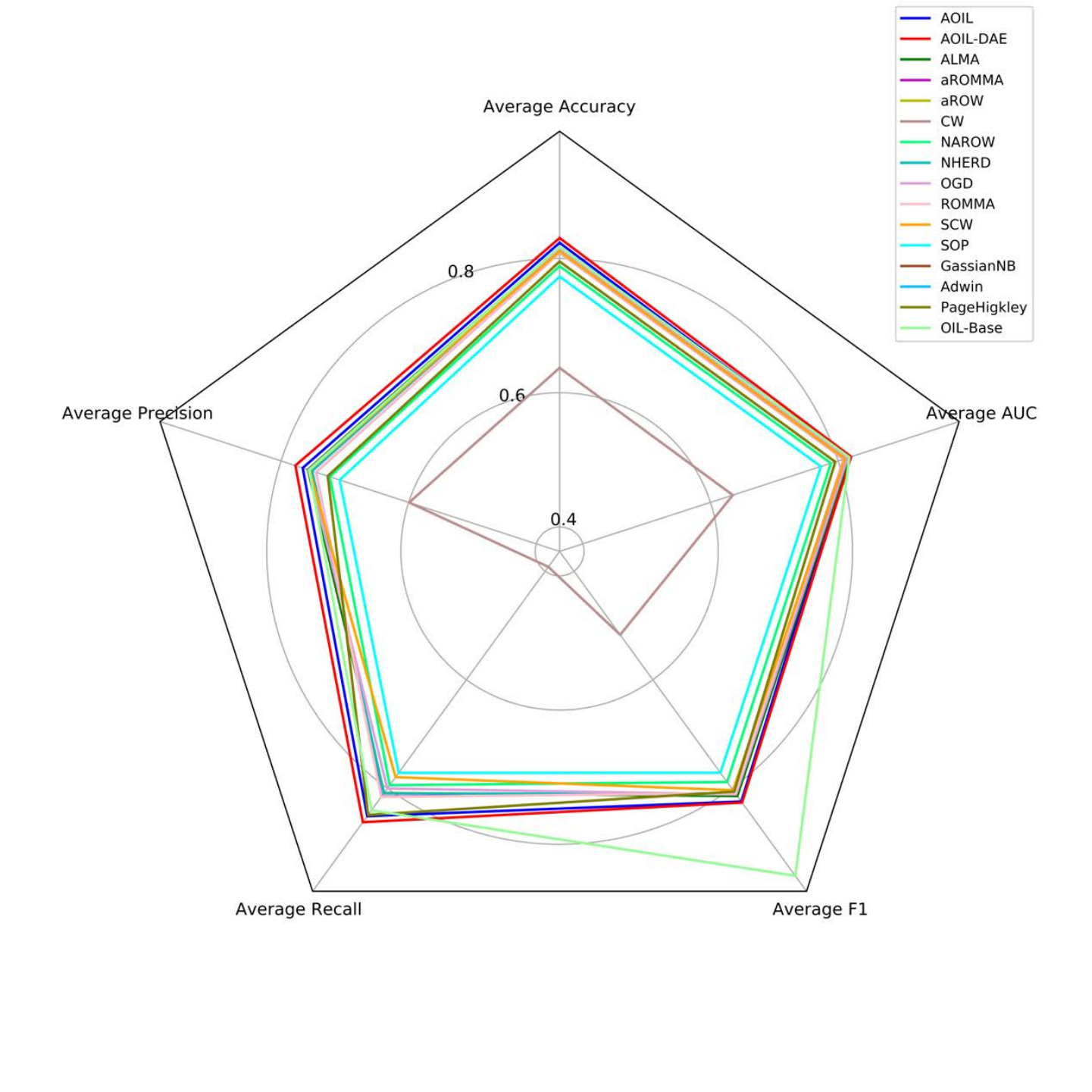}
	}
	\quad
	\subfigure[susy(noise)]{
	\includegraphics[width=7cm]{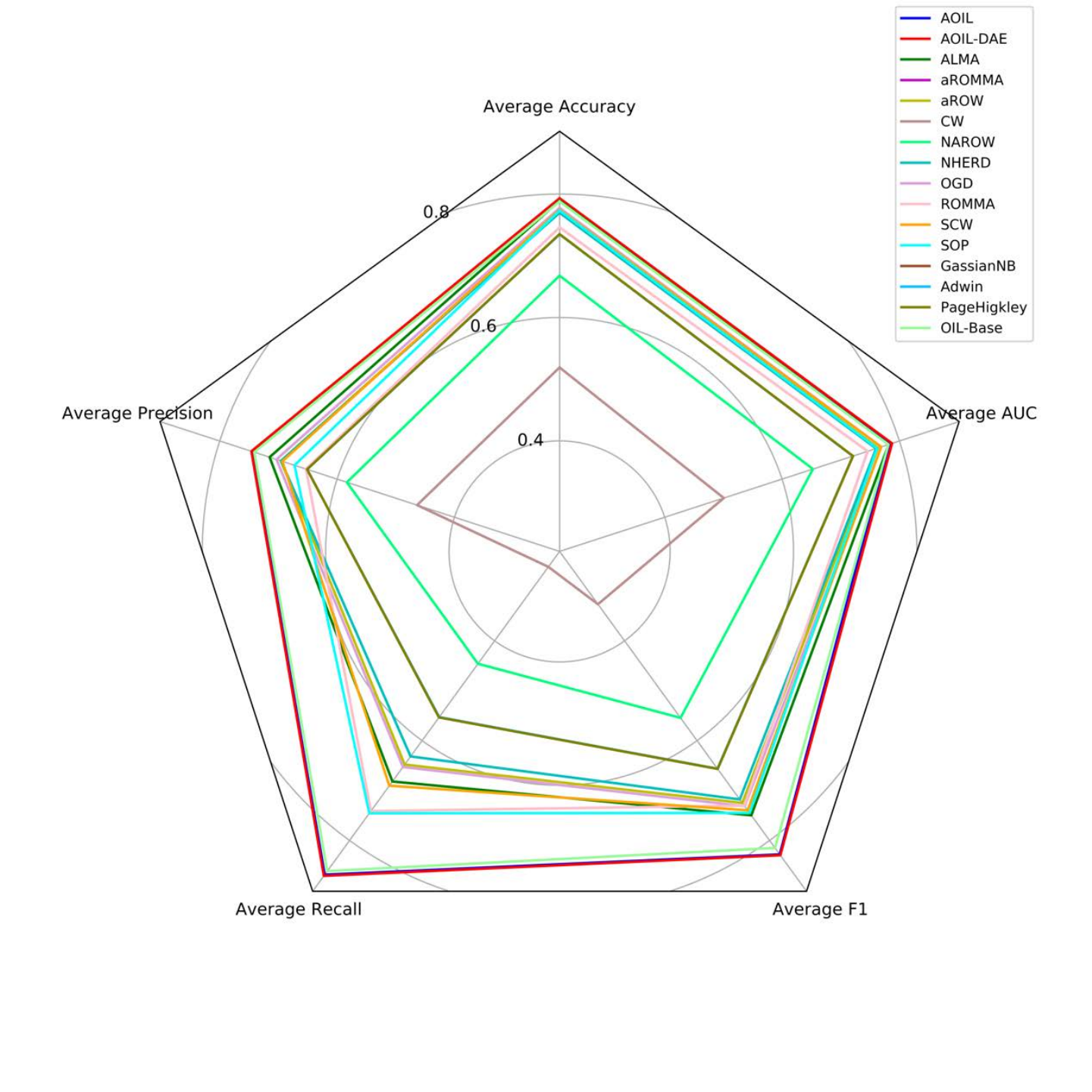}

	}
	\caption{Different algorithms’ performance on different evaluation indexes on the noisy data.}
\end{figure}
\begin{table}[!htbp]
	\centering
	\caption{The comparative results of different algorithms on different noisy datasets.}
	\label{tb2}
	\begin{tabular}{cccccccc}\hline
		\multirow{3}{*}{\textbf{method}} &\multicolumn{7}{c}{Average Accuracy}\\
		\cline{2-8}
		 &weather  &sea  &hyperplane  &occupancy  &kddcup  &hepmass  &susy\\
		 &(noise)&(noise)&(noise)&(noise)&(noise)&(noise)&(noise)\\\hline
AOIL-DAE	     &76.68\%	&91.03\%	&90.58\%	&98.11\%	&99.62\%	&83.05\%	&79.32\%\\
AOIL	          & 76.01\%	&90.87\%	&88.73\%	&96.52\%	&99.52\%	&82.34\%	&79.23\%\\
ALMA	          & 75.31\%	&76.40\%	&66.41\%	&95.93\%	&99.48\%	&81.69\%	&78.90\%\\
aROMMA	     &74.94\%	&33.75\%	&43.62\%	&94.52\%	&99.39\%	&80.87\%	&74.60\%\\
aROW	&73.17\%	&76.51\%	&53.35\%	&95.70\%	&67.95\%	&81.30\%	&77.19\%\\
CW	&65.33\%	&32.88\%	&41.58\%	&95.37\%	&68.19\%	&63.70\%	&51.90\%\\
NAROW	&70.03\%	&71.47\%	&55.99\%	&91.52\%	&68.09\%	&78.86\%	&66.78\%\\
NHERD	&73.09\%	&76.54\%	&53.27\%	&95.29\%	&67.84\%	&81.24\%	&76.97\%\\
OGD	&70.63\%	&76.54\%	&66.41\%	&95.07\%	&99.28\%	&81.49\%	&77.76\%\\
ROMMA	&74.85\%	&33.77\%	&43.67\%	&94.51\%	&99.34\%	&80.90\%	&74.60\%\\
SCW	&72.95\%	&76.31\%	&53.04\%	&95.80\%	&68.06\%	&81.08\%	&77.47\%\\
SOP	&55.84\%	&78.17\%	&70.80\%	&94.21\%	&67.96\%	&77.30\%	&76.33\%\\
GassianNB	     &66.64\%	&87.84\%	&84.46\%	&96.51\%	&98.75\%	&79.55\%	&73.50\%\\
Adwin	&68.56\%	&90.40\%	&88.93\%	&91.05\%	&98.75\%	&79.55\%	&73.47\%\\
PageHigkley	&66.64\%	&87.53\%	&84.46\%	&96.51\%	&98.75\%	&79.55\%	&73.50\%\\
OIL-Base	     &72.03\%	&86.79\%	&86.76\%	&82.34\%	&98.47\%	&81.75\%	&78.80\%\\
\hline
	\end{tabular}
\end{table}
\begin{table}[!htbp]
	\centering
	\begin{tabular}{cccccccc}\hline
		\multirow{2}{*}{\textbf{method}} &\multicolumn{7}{c}{Average Precision}\\
		\cline{2-8}
		&weather  &sea  &hyperplane  &occupancy  &kddcup  &hepmass  &susy\\
		 &(noise)&(noise)&(noise)&(noise)&(noise)&(noise)&(noise)\\\hline
AOIL-DAE	&0.7899	&0.8982	&0.8695	&0.9874	&0.9972	&0.7775	&0.7460\\
AOIL	     &0.7807	&0.8855	&0.8472	&0.9523	&0.9801	&0.7660	&0.7452\\
ALMA	     & 0.4965	&0.7881	&0.6684	&0.8471	&0.9969	&0.7578	&0.7150\\
aROMMA    & 0.5124	&0.7302	&0.6039	&0.7991	&0.9949	&0.7456	&0.6530\\
aROW	     & 0.4617	&0.7859	&0.5909	&0.8387	&0.8010	&0.7525	&0.6958\\
CW	     & 0.3288	&0.7221	&0.5903	&0.8273	&0.8011	&0.5993	&0.4637\\
NAROW     & 0.4257	&0.7146	&0.5917	&0.6979	&0.8009	&0.7236	&0.5833\\
NHERD	     & 0.4578	&0.7839	&0.5910	&0.8240	&0.8010	&0.7515	&0.6954\\
OGD	     & 0.3988	&0.7859	&0.6679	&0.8176	&0.9976	&0.7567	&0.7033\\
ROMMA	     & 0.5114	&0.7302	&0.6040	&0.7987	&0.9944	&0.7459	&0.6530\\
SCW	     & 0.4633	&0.7898	&0.5912	&0.8425	&0.8010	&0.7554	&0.6932\\
SOP	     & 0.3305	&0.8293	&0.7136	&0.7872	&0.8008	&0.7081	&0.6726\\
GassianNB	& 0.4180	&0.8598	&0.7913	&0.8601	&0.9909	&0.7268	&0.6512\\ 
Adwin	     & 0.4333	&0.8863	&0.8465	&0.6433	&0.9909	&0.7268	&0.6510\\
PageHigkley	     &0.4180	&0.8531	&0.7913	&0.8601	&0.9909	&0.7268	&0.6512\\
OIL-Base	& 0.7327	&0.7443	&0.8231	&0.8236	&0.9352	&0.7551	&0.7411\\
\hline
	\end{tabular}
\end{table}
\begin{table}[!htbp]
	\centering
	\begin{tabular}{cccccccc}\hline
		\multirow{2}{*}{\textbf{method}} &\multicolumn{7}{c}{Average Recall}\\
		\cline{2-8}
		&weather  &sea  &hyperplane  &occupancy  &kddcup  &hepmass  &susy\\
		 &(noise)&(noise)&(noise)&(noise)&(noise)&(noise)&(noise)\\\hline
AOIL-DAE	&0.8967	&0.9634	&0.8947	&0.9908	&0.9970	&0.8623	&0.8708\\
AOIL	     &0.8964	&0.9432	&0.8746	&0.9762	&0.9901	&0.8512	&0.8685\\
ALMA	     &0.4771	&0.8999	&0.7843	&0.9072	&0.9957	&0.8084	&0.6818\\
aROMMA	&0.6372	&0.0808	&0.0606	&0.8365	&0.9968	&0.8152	&0.7411\\
aROW	&0.4230	&0.9080	&0.6846	&0.9023	&0.7977	&0.8087	&0.6479\\
CW	&0.039	&0.0780	&0.0420	&0.8891	&0.8013	&0.3920	&0.2518\\
NAROW	&0.4126	&0.9999	&0.8222	&0.7866	&0.8001	&0.7941	&0.4459\\
NHERD	&0.4001	&0.9140	&0.6790	&0.9089	&0.7957	&0.8093	&0.6316\\
OGD	&0.1624	&0.9087	&0.7880	&0.8688	&0.9918	&0.8002	&0.6528\\
ROMMA	&0.6380	&0.0810	&0.0607	&0.8359	&0.9967	&0.8158	&0.7411\\
SCW	&0.4545	&0.8935	&0.6640	&0.9007	&0.7994	&0.7793	&0.6902\\
SOP	&0.3342	&0.8420	&0.7501	&0.8694	&0.7982	&0.7714	&0.7455\\
GassianNB	     &0.6399	&0.9436	&0.8454	&0.9392	&0.9220	&0.8490	&0.5534\\
Adwin	          &0.6439	&0.9563	&0.8893	&0.6386	&0.9920	&0.8490	&0.5534\\
PageHigkley	&0.6399	&0.9512	&0.8454	&0.9392	&0.9920	&0.8490	&0.5534\\
OIL-Base	     &0.8457	&0.7757	&0.8510	&0.9860	&0.9443	&0.8404	&0.8613\\
\hline
	\end{tabular}
\end{table}\begin{table}[!htbp]
	\centering
	\begin{tabular}{cccccccc}\hline
		\multirow{2}{*}{\textbf{method}} &\multicolumn{7}{c}{Average F1}\\
		\cline{2-8}
		 &weather  &sea  &hyperplane  &occupancy  &kddcup  &hepmass  &susy\\
		 &(noise)&(noise)&(noise)&(noise)&(noise)&(noise)&(noise)\\\hline
AOIL-DAE	&0.8502	&0.9317	&0.8976	&0.9879	&0.9967	&0.8262	&0.8295\\
AOIL   	&0.8368	&0.9071	&0.8746	&0.9721	&0.9833	&0.8242	&0.8279\\
ALMA	      &0.5605	&0.8449	&0.7340	&0.9090	&0.9967	&0.8141	&0.7494\\
aROMMA	 &0.6266	&0.1485	&0.1128	&0.8724	&0.9961	&0.8086	&0.7300\\
aROW	&0.5099	&0.8467	&0.6343	&0.9038	&0.7994	&0.8110	&0.7247\\
CW	&0.0690	&0.1439	&0.079	&0.8958	&0.8013	&0.5172	&0.3267\\
NAROW	&0.4760	&0.8335	&0.6882	&0.8060	&0.8005	&0.7883	&0.5543\\
NHERD	&0.4952	&0.8477	&0.6320	&0.8964	&0.7984	&0.8105	&0.7177\\
OGD	&0.2673	&0.8470	&0.7349	&0.8876	&0.9955	&0.8109	&0.7312\\
ROMMA	&0.6260	&0.1492	&0.1129	&0.8721	&0.9959	&0.8090	&0.7300\\
SCW	&0.5257    &0.8435	&0.6256	&0.9056	&0.8003	&0.8033	&0.7395\\
SOP	&0.3330	&0.8464	&0.7522	&0.8706	&0.7995	&0.7712	&0.7448\\
GassianNB	     &0.5463	&0.9072	&0.8449	&0.9202	&0.9922	&0.8060	&0.6566\\
Adwin	          & 0.5625	&0.9262	&0.8894	&0.7560	&0.9922	&0.8060	&0.6566\\
PageHigkley	&0.5463	&0.9058	&0.8449	&0.9202	&0.9922	&0.8060	&0.6566\\
OIL-Base	     &0.8180	&0.8124	&0.8652	&0.8978	&0.9611	&0.8216	&0.8149\\
\hline
	\end{tabular}
\end{table}\begin{table}[!htbp]
	\centering
	\begin{tabular}{cccccccc}\hline
		\multirow{2}{*}{\textbf{method}} &\multicolumn{7}{c}{Average AUC}\\
		\cline{2-8}
		 &weather  &sea  &hyperplane  &occupancy  &kddcup  &hepmass  &susy\\
		 &(noise)&(noise)&(noise)&(noise)&(noise)&(noise)&(noise)\\\hline
AOIL-DAE	&0.7302	&0.8934	&0.8958	&0.9737	&0.9942	&0.8203	&0.7874\\
AOIL	     &0.7011	&0.8882	&0.8763	&0.9544	&0.9846	&0.8152	&0.7825\\
ALMA	     &0.6830	&0.6619	&0.6374	&0.9408	&0.9933	&0.8168	&0.7817\\
aROMMA	&0.7209	&0.5303	&0.5197	&0.9065	&0.9895	&0.8087	&0.7457\\
aROW	&0.6533	&0.6577	&0.5000	&0.9375	&0.5016	&0.8130	&0.7634\\
CW	&0.4974	&0.5165	&0.4988	&0.9307	&0.5020	&0.6350	&0.5007\\
NAROW	&0.6272	&0.5004	&0.5016	&0.8694	&0.5013	&0.7886	&0.6526\\
NHERD	&0.6469	&0.6538	&0.5002	&0.9373	&0.5016	&0.8123	&0.7603\\
OGD	&0.5682	&0.6578	&0.6366	&0.9216	&0.9943	&0.8148	&0.7691\\
ROMMA	&0.7205	&0.5303	&0.5198	&0.9062	&0.9885	&0.8090	&0.7456\\
SCW	&0.6597	&0.6652	&0.5007	&0.9376	&0.5016	&0.8105	&0.7689\\
SOP	&0.5014	&0.7364	&0.6987	&0.9162	&0.5008	&0.7730	&0.7621\\
GassianNB	     &0.6592	&0.8553	&0.8446	&0.9557	&0.9809	&0.7955	&0.7209\\
Adwin	&0.6743	&0.8855	&0.8893	&0.8165	&0.9809	&0.7955	&0.7206\\
PageHigkley	&0.6592	&0.8485	&0.8446	&0.9557	&0.9809	&0.7955	&0.7209\\
OIL-Base	     &0.6057	&0.8487	&0.8676	&0.6046	&0.9695	&0.8175	&0.7813\\
\hline
	\end{tabular}
\end{table}
\begin{figure}[!htbp]
	\centering
	\subfigure[Average Accuracy]{
		\includegraphics[width=4.5cm]{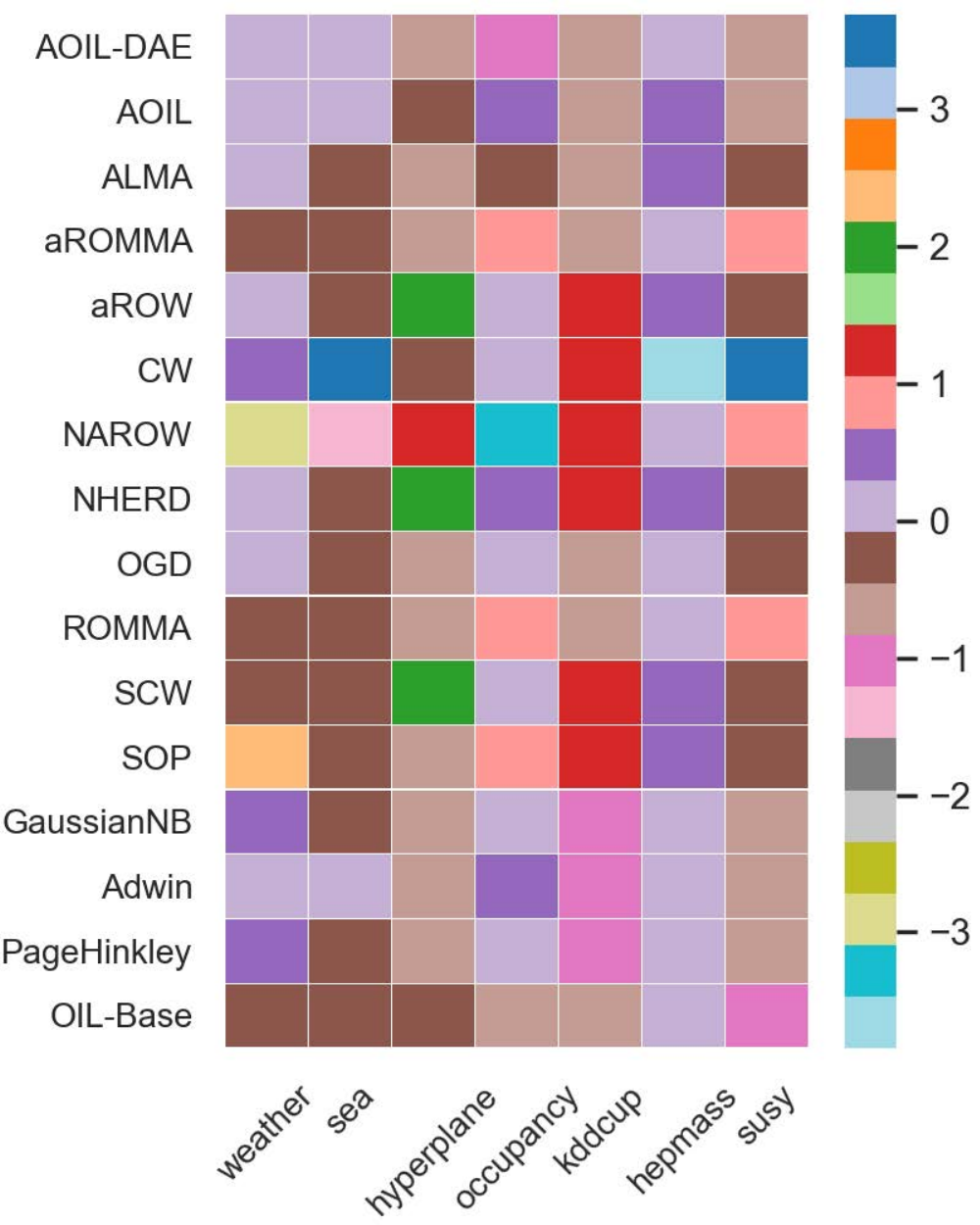}
	}
	\quad
	\subfigure[Average Precision]{
		\includegraphics[width=4.5cm]{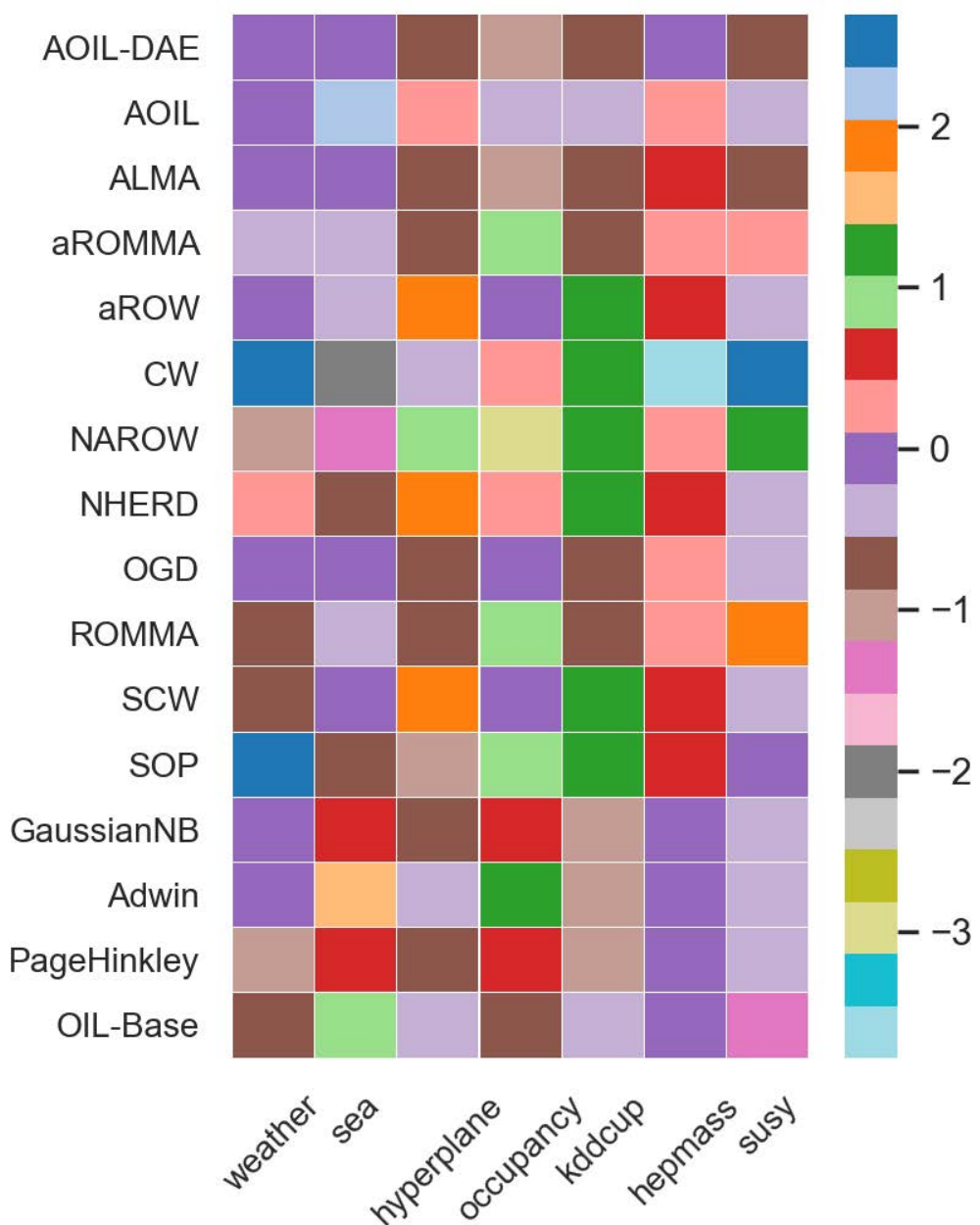}
	}
	\quad
	\subfigure[ Average Recall]{
		\includegraphics[width=4.5cm]{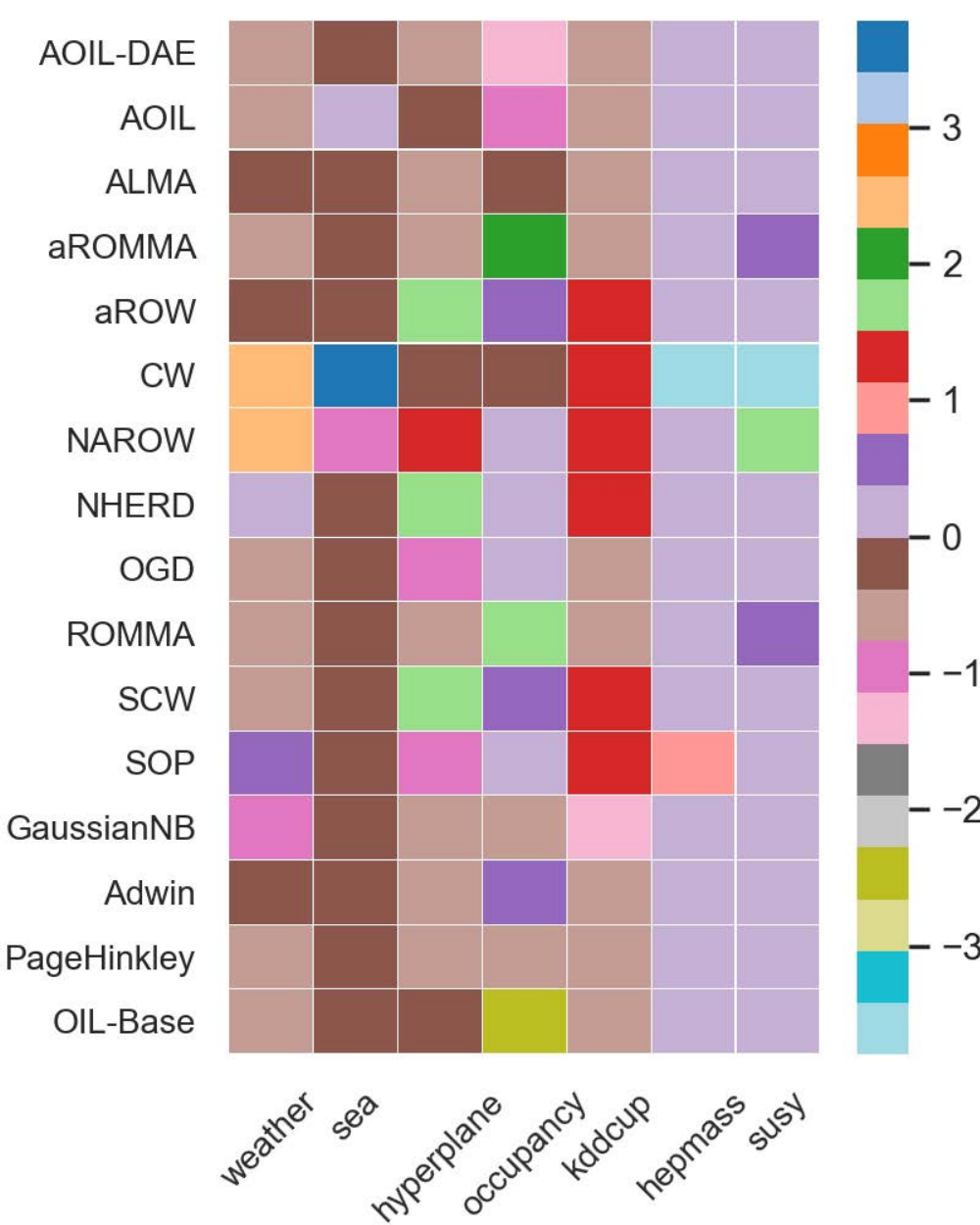}
	}
	\subfigure[ Average F1]{
		\includegraphics[width=4.5cm]{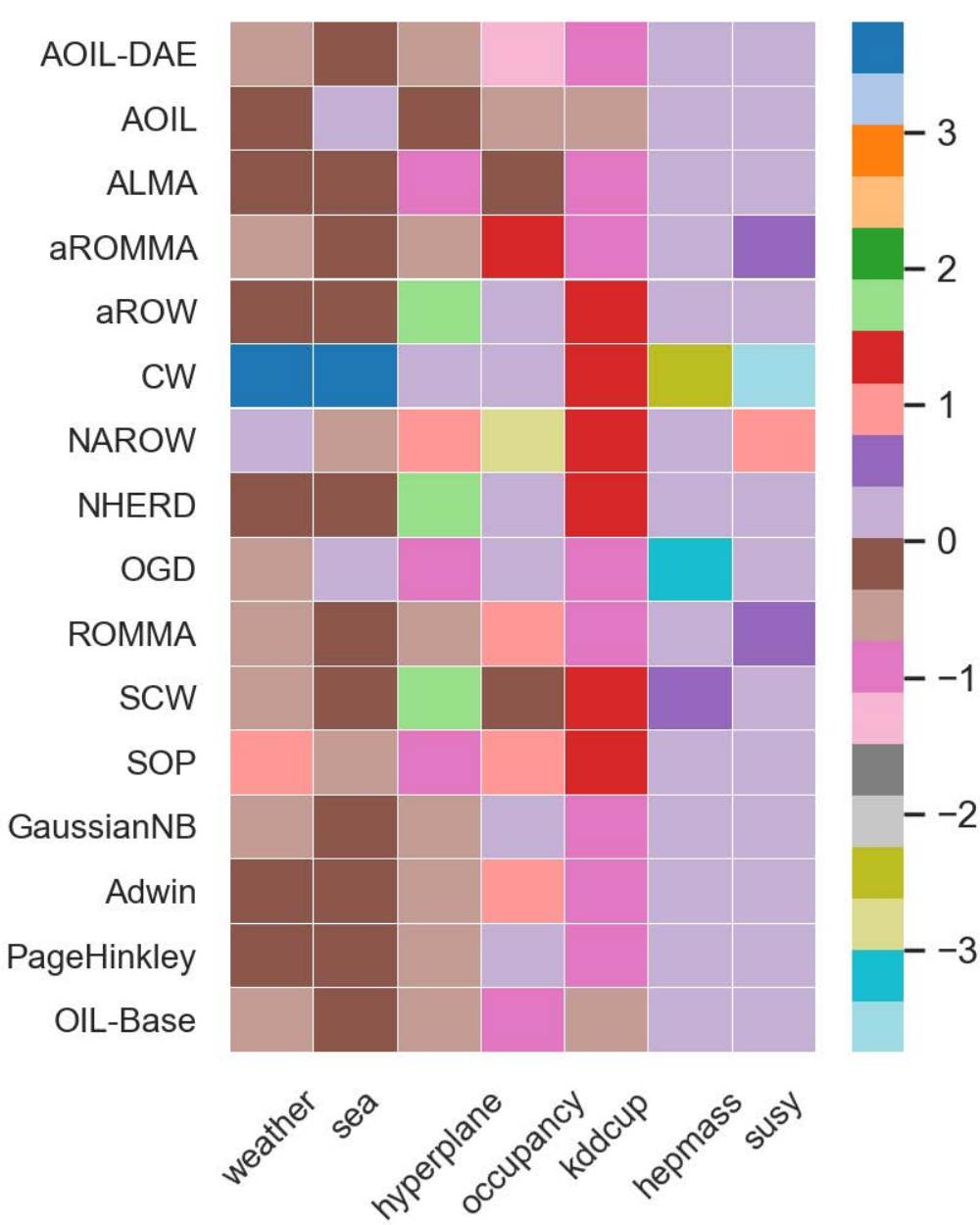}
	}
	\quad
	\subfigure[Average AUC]{
	\includegraphics[width=4.5cm]{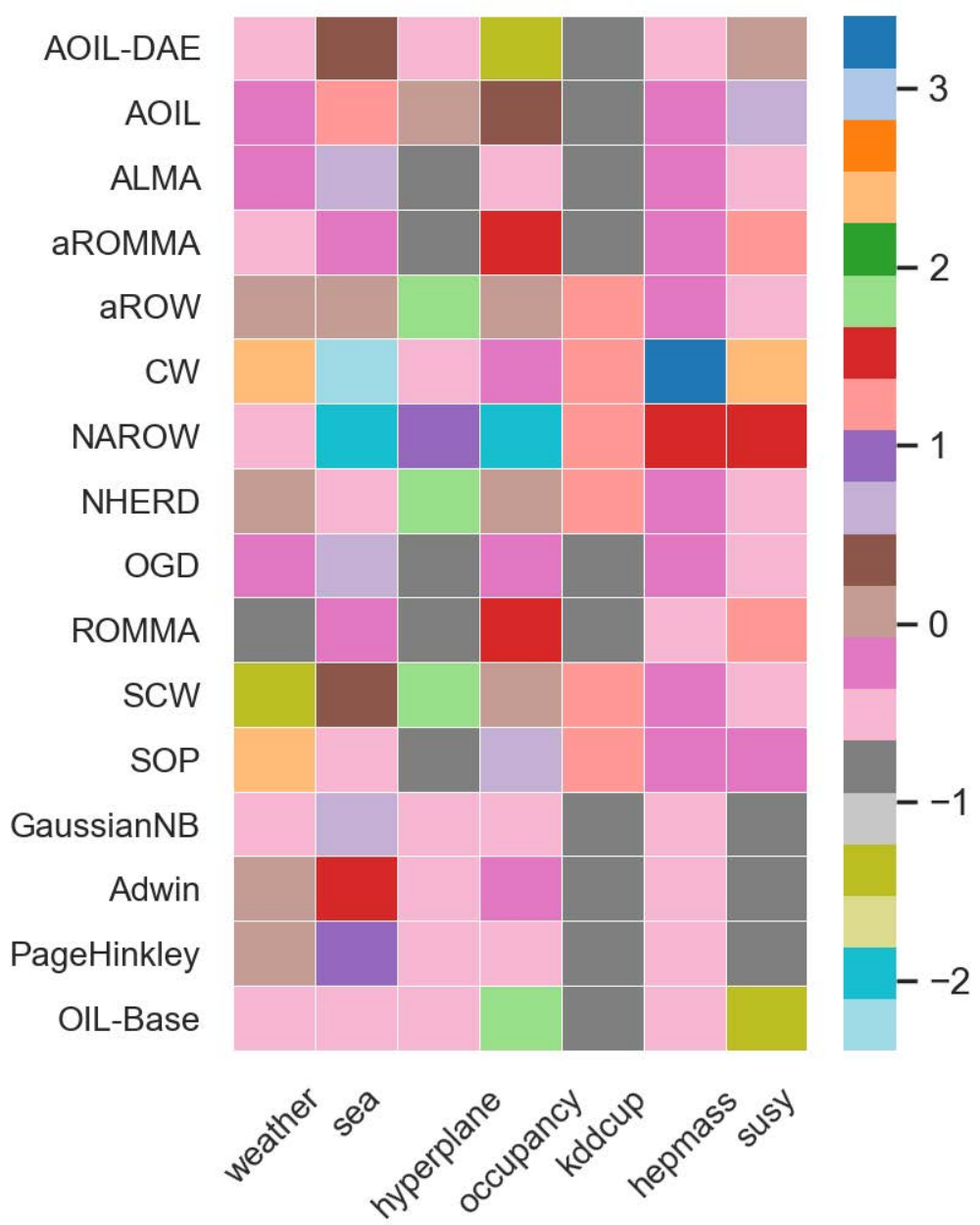}

	}
	\caption{These algorithms’ heat-map for each evaluation criteria after adding noise to streaming datasets. The horizontal axis represents different data sets, and the vertical axis represents different algorithms.}
\end{figure}

We plotted heat maps as shown in Fig.10 for illustrating the increase and decrease of each evaluation criteria after adding noise to these streaming datasets. The horizontal axis represents different data sets, and the vertical axis represents different algorithms. When adding noise to the data, of course, we hope that the performance change of the algorithm is small. In the heat maps, we use the different color to represent the changes in predictive performance. From the color in Fig.10, we can see that our algorithm AOIL is less affected by noise, which shows that the AOIL has better robustness.
\begin{figure}[!htbp]
	\centering
	\subfigure[weather(noise)]{
		\includegraphics[width=7cm]{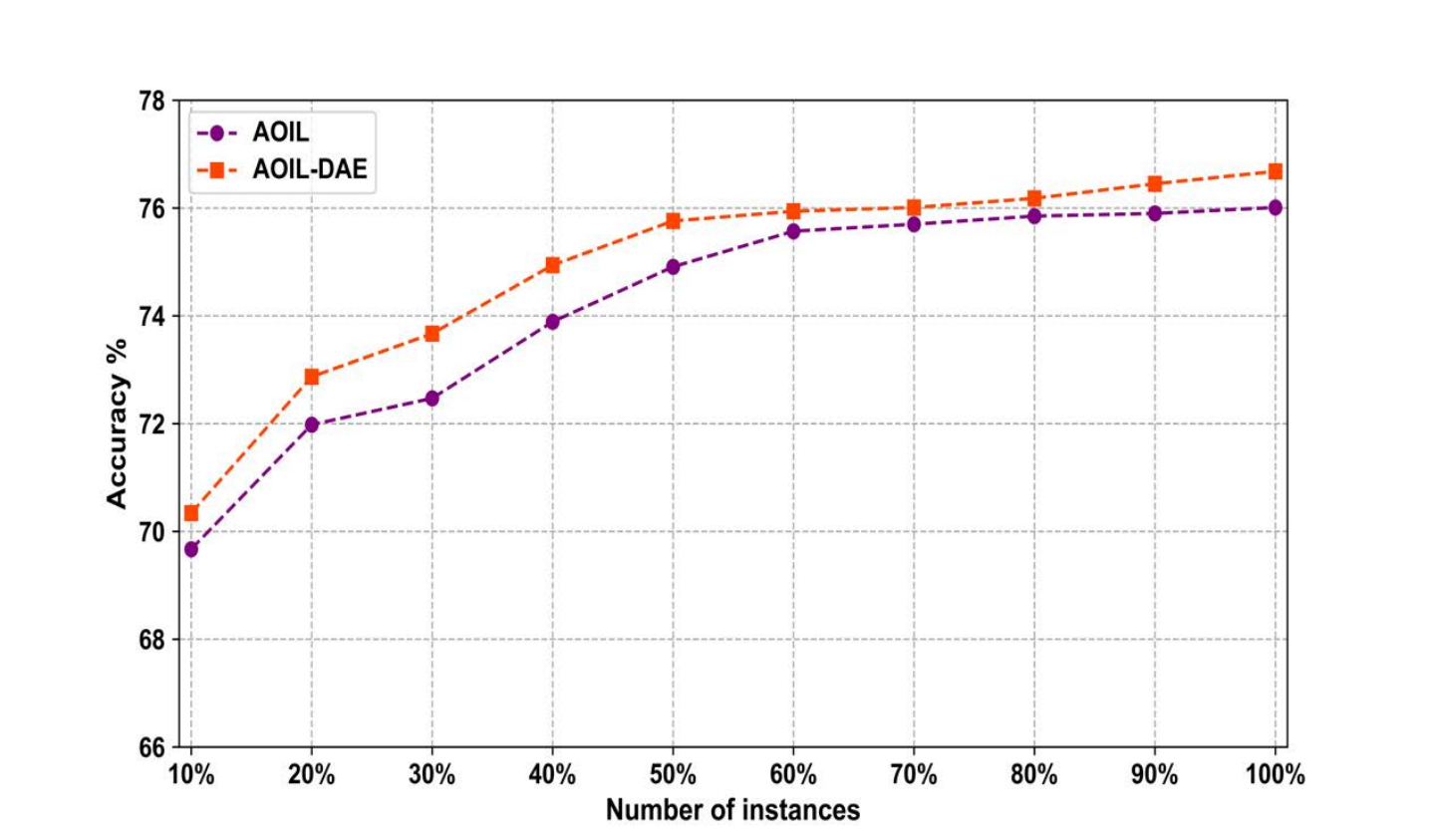}
	}
	\quad
	\subfigure[sea(noise)]{
		\includegraphics[width=7cm]{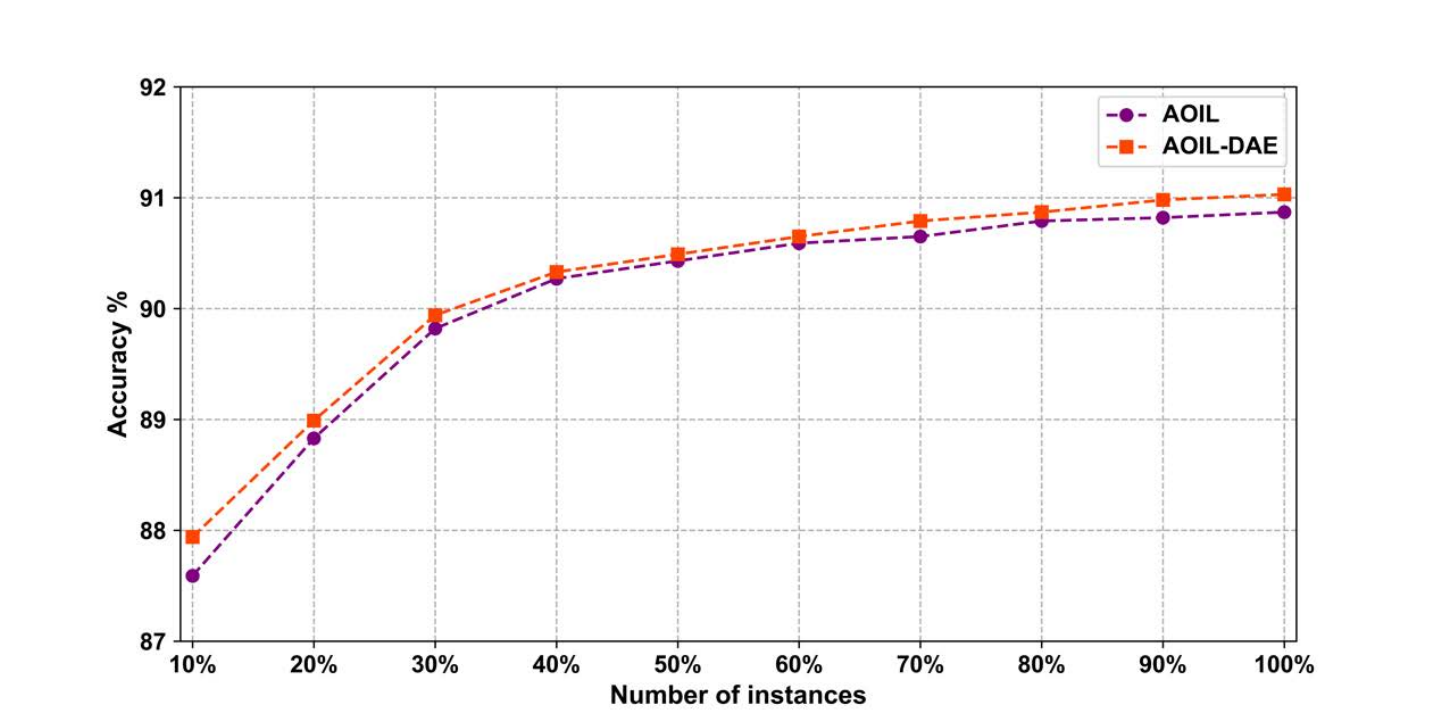}
	}
	\quad
	\subfigure[hyperplane(noise)]{
		\includegraphics[width=7cm]{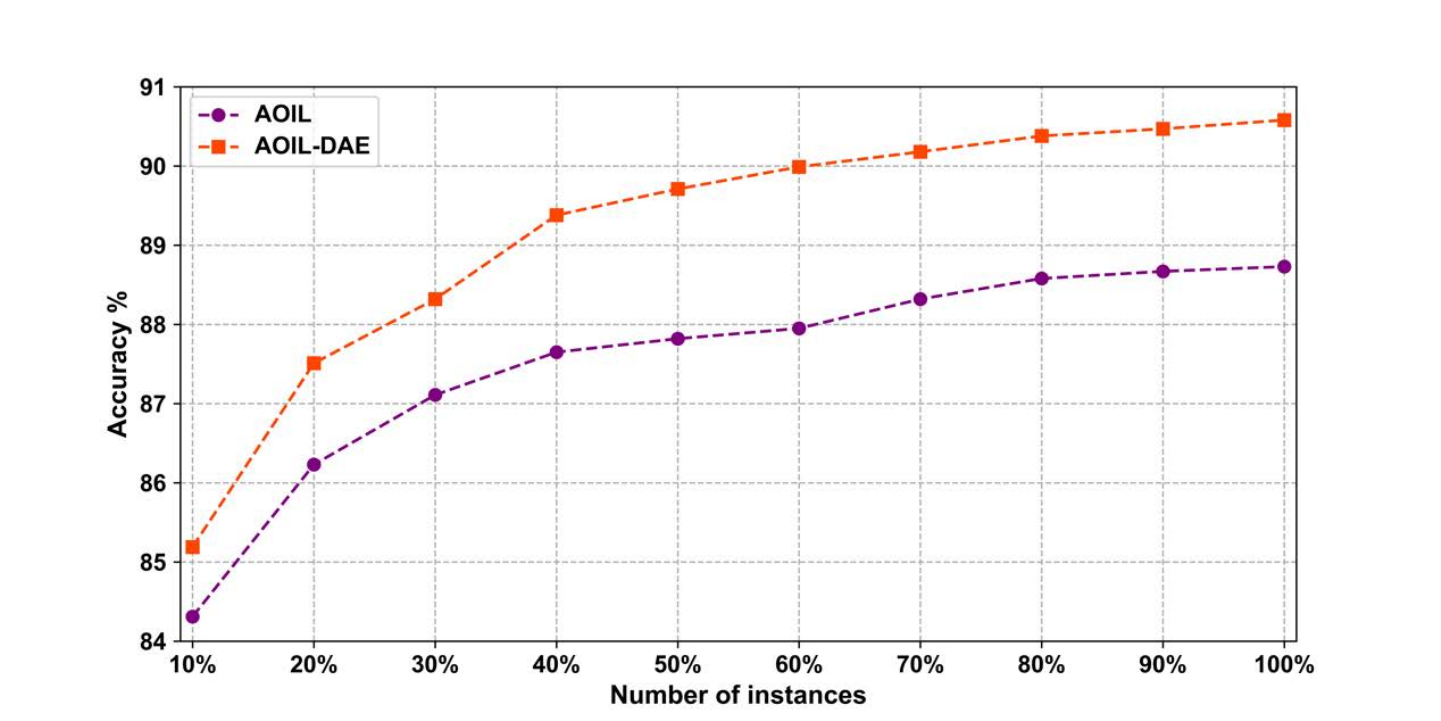}
	}
	\subfigure[occupancy(noise)]{
		\includegraphics[width=7cm]{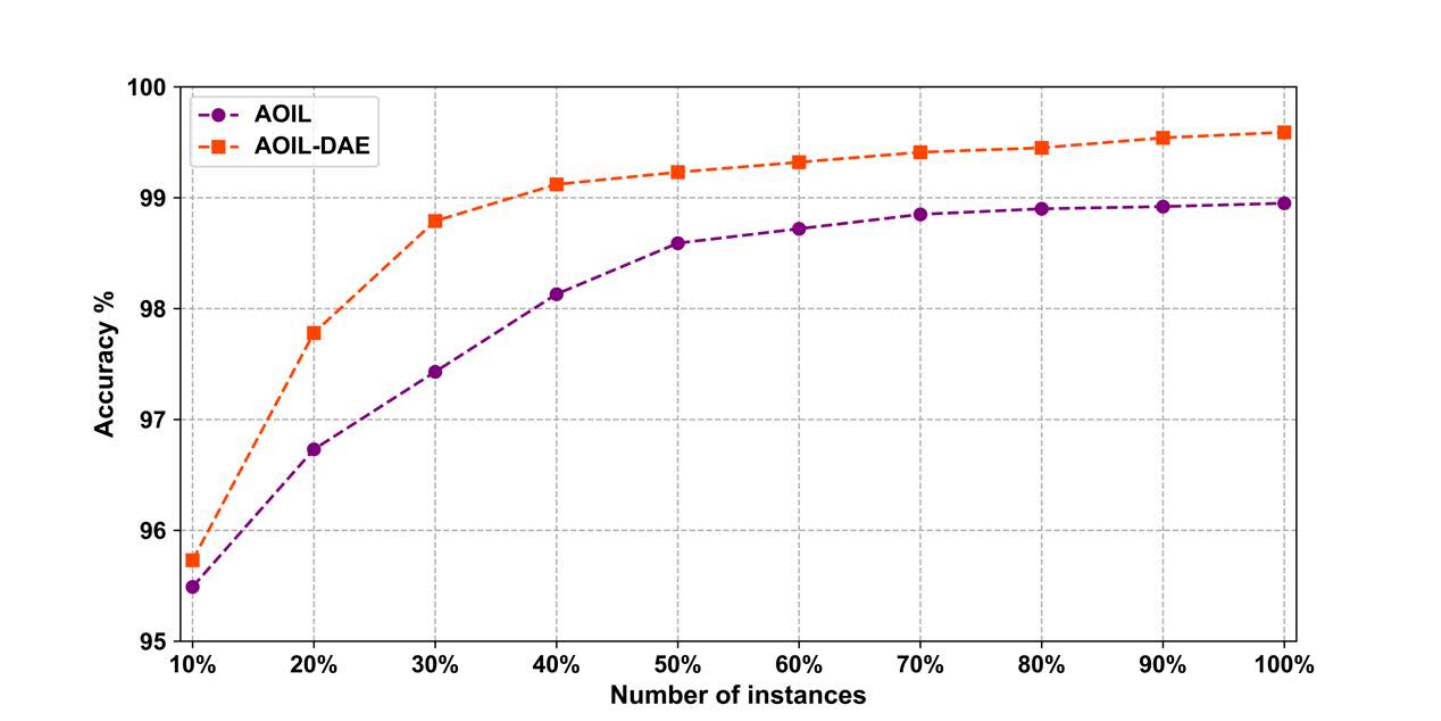}
	}
	\quad
	\subfigure[kddcup(noise)]{
		\includegraphics[width=7cm]{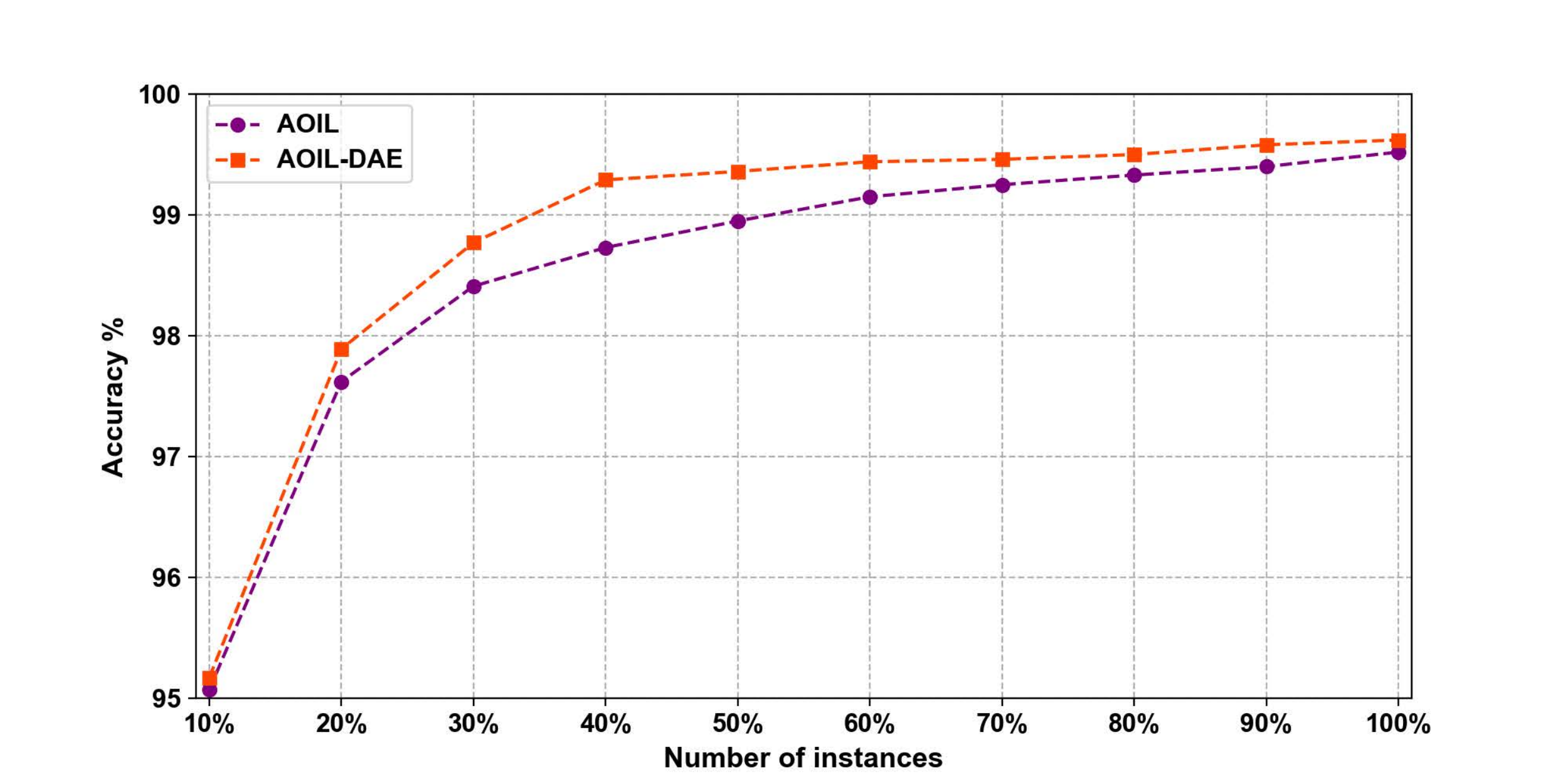}
	}
	\quad
	\subfigure[hepmass(noise)]{
		\includegraphics[width=7cm]{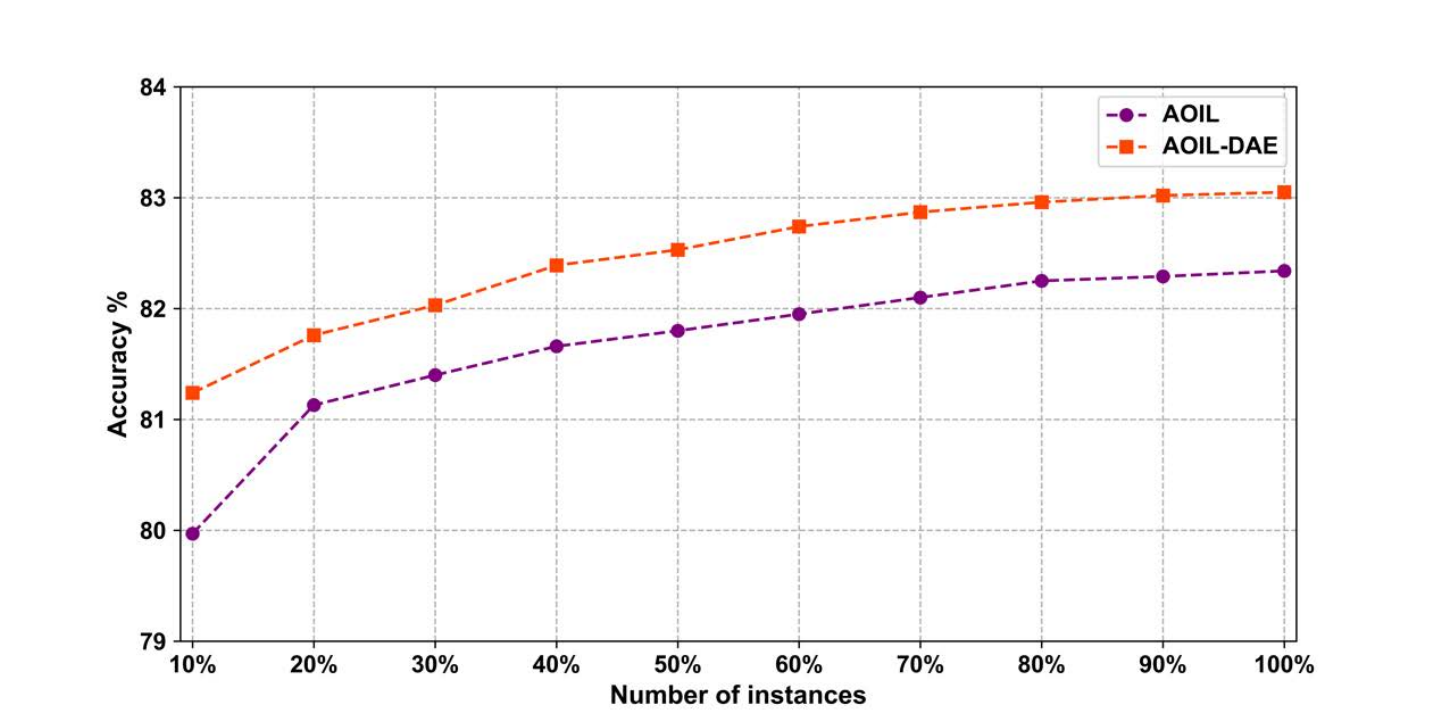}
	}
	\quad
	\subfigure[susy(noise)]{
	\includegraphics[width=9cm]{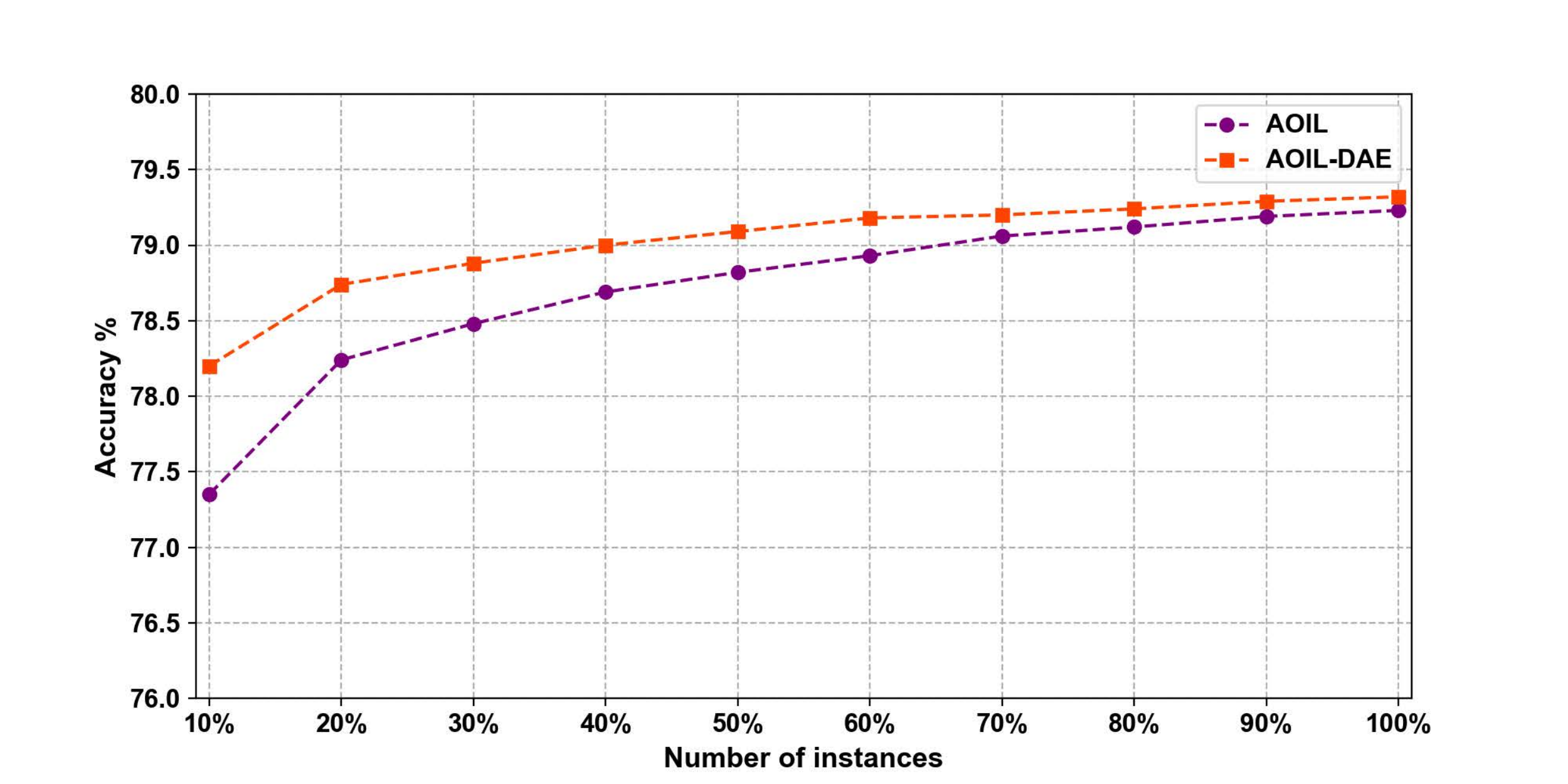}

	}
	\caption{The accuracies of AOIL and AOIL-DAE algorithm on different noisy datasets.}
\end{figure}

 In order to make the comparison more obvious, we tend to compare the two algorithms, the AOIL-DAE and AOIL separately, as we can see in Fig.11, obviously, we can see that after adding noise, AOIL-DAE performs better than AOIL. In general, after adding noise, AOIL-DAE and AOIL have not changed much for each evaluation criteria, and the AOIL-DAE algorithm has better robustness than AOIL. So when there is a lot of noise in the data, we recommend AOIL-DAE algorithm.

\section{Conclusion and Future Work}

In this paper, we propose an adaptive online incremental learning algorithm based on an auto-encoder with the memory module. As we mentioned before, online incremental learning mainly faces three problems. The first problem is catastrophic forgetting. Some algorithms blindly pursue to learn new knowledge and ignore the learned knowledge of the algorithm, which is against the idea of lifelong learning. The second is the problem of concept drift in data streaming. Because the data distribution in the data stream would change at any time, it is necessary to detect the concept drift in time and trigger the update mechanism to improve the predictive accuracies of the algorithm. The last problem is also an issue that is easy to be ignored in the online learning field, that is, latent representation learning. Because the knowledge we learn is through the hidden layer features, it is particularly important to lay a good foundation, that is, to learn a good implicit representation and extract useful hidden layer information. Here, our goal is to meet these three demands in a unified framework. Along this route, we develop an adaptive online incremental learning model (AOIL), which can effectively extract the input information, accurately detect the existence of concept drift, and trigger the update mechanism. At the same time, in the feature extraction layer, different hidden layers play different roles. It can extract the common and private features of the input data at the same time, effectively reducing the problem of catastrophic forgetting. Then, the final fusion features are obtained using self-attention mechanism for different features. 

Moreover, to validate the effectiveness of our algorithm, we carry out experiments on both stational datasets and non-stational datasets. In the experiments, we use five different evaluation indexes to compare our algorithm with other baseline algorithms. The numerical results demonstrate the advantage of our algorithm. In addition, as we can see from the accuracies in different learning stages, we can see that the AOIL algorithm could effectively reduce the catastrophic forgetting problem. Finally, we propose the AOIL-DAE algorithm to further improve the robustness of the AOIL algorithm.

Nevertheless, it is admitted that AOIL and AOIL-DAE only suitable for processing a single task for data stream at present. We did not take the transfer learning problem across several streaming processes problem into account.

In future work, we would consider the online incremental learning problem for multiple related data streams. Because there is widespread multiple data streaming in many practical scenarios, it can be regarded as multi-task data streams. In building data streams modeling on a single task, it would be more effective to consider the correlation between multi-task data streams instead of dealing with them separately.

\bibliography{bibfile}

\end{document}